\documentclass{article}
\usepackage{subfiles}
\usepackage[margin=1in]{geometry}
\usepackage{cite}
\usepackage{color}
\definecolor{darkblue}{rgb}{0.1,0.1,1.0}

\DeclareRobustCommand{\Sec}[1]{Section~\ref{sec:#1}}
\DeclareRobustCommand{\Tab}[1]{Table~\ref{tab:#1}}

\DeclareRobustCommand{\Fig}[1]{Figure~\ref{fig:#1}}

\DeclareRobustCommand{\Eq}[1]{Equation~(\ref{eq:#1})}

\usepackage{sidecap}
\sidecaptionvpos{figure}{t}
\usepackage{microtype}
\usepackage{graphicx}
\usepackage{booktabs}

\usepackage{amsmath}
\usepackage{amssymb}
\usepackage{amsthm}
\usepackage[colorlinks=true,urlcolor=blue,anchorcolor=blue,citecolor=blue,filecolor=blue,linkcolor=blue,menucolor=blue,linktocpage=true]{hyperref}
\usepackage{graphicx}
\usepackage[export]{adjustbox}
\usepackage{subfig}
\usepackage{float}

\newtheorem{thm}{Theorem}
\newtheorem{lemma}{Lemma}
\newtheorem{definition}{Definition}

\newtheorem{conjecture}{Conjecture}

\newcommand{\es}[2] {\begin{equation} \label{#1} \begin{split} #2 \end{split} \end{equation}}

\newcommand{\lexpp}[1]{\mathbb{E}_{#1}\left[}
\newcommand{\lvar}[1]{\text{Var}_{#1}\left[}
\newcommand{\rexp}{\right]}

\newcommand{\DL}{\textrm{DL}}

\usepackage[dvipsnames]{xcolor}
\definecolor{darkgreen}{HTML}{008000}
\usepackage{color}
\definecolor{purple}{rgb}{0.5,0,0.5}

\usepackage{authblk}
\begin{document}

\title{Asymptotics of Wide Convolutional Neural Networks}

\author{Anders Andreassen\thanks{Corresponding Author: ajandreassen@google.com} }
\author{Ethan Dyer\thanks{edyer@google.com}}
\affil{Google, Mountain View, CA 94043, USA}

\vskip 0.3in

\maketitle

\begin{abstract}
Wide neural networks have proven to be a rich class of architectures for both theory and practice. 
Motivated by the observation that finite width convolutional networks appear to outperform infinite width networks, we study scaling laws for wide CNNs and networks with skip connections.
Following the approach of \cite{dyer2020asymptotics}, we present a simple diagrammatic recipe to derive the asymptotic width dependence for many quantities of interest.
These scaling relationships provide a solvable description for the training dynamics of wide convolutional networks.
We test these relations across a broad range of architectures. In particular, we find that the difference in performance between finite and infinite width models vanishes at a definite rate with respect to model width. Nonetheless, this relation is consistent with finite width models generalizing either better or worse than their infinite width counterparts, and we provide examples where the relative performance depends on the optimization details. 
\end{abstract}
\section{Introduction}

Deep neural networks continue to achieve remarkable performance on a diverse range of machine learning tasks, however detailed understanding remains elusive.
One of the most promising routes towards understanding is to study very wide neural networks. Wide networks strike an attractive balance between performance \cite{arora2020harnessing, neyshabur2018the} and analytic control \cite{NTK,2019arXiv190206720L}. Furthermore understanding the performance of networks as the number of parameters is increased is at the heart of the generalization paradox -- the observation that over-parameterized deep networks do not over-fit.

In \cite{NTK} the authors showed that the dynamics of infinitely wide fully connected (FC) neural networks trained under gradient flow simplifies dramatically, and in the limit of infinite width \cite{2019arXiv190206720L} argued that training a deep network is equivalent to training a linear random features model. For models trained with mean squared error (MSE) loss, this infinite width, linear evolution can be written as
\es{linearevol}{
\frac{df(x)}{dt}=-\sum_{x_{a}\in\mathcal{D}_{\textrm{train}}}\Theta(x,x_{a})\left(f(x_{a})-y_{a}\right)\,,
}
where $f(x)$ is the network output on example $x$, $\mathcal{D}_{\textrm{train}}$ is the training dataset, and $\Theta(x,x')=\sum_{\mu}\frac{\partial f(x)}{\partial\theta_{\mu}}\frac{\partial f(x')}{\partial\theta_{\mu}}$ is the neural tangent kernel (NTK).

In \cite{dyer2020asymptotics,huang2019dynamics} this simplified infinite width evolution was extended to take into account corrections from finite width, giving a controlled prediction for the evolution of wide fully connected networks trained via stochastic gradient descent (SGD).

This formalism has the promise of explaining the training dynamics and predictions of wide neural networks. Empirically, however, there are several results that show convolutional neural networks (CNNs) exhibiting different behavior from fully connected networks (FCs) that have yet to be understood.
First, for CNNs, but not FCs, there is evidence that finite width networks trained via SGD outperform their infinite width counterparts \cite{Arora2019OnEC, novak2019bayesian, lee2020finite}.
Paradoxically, despite the drop in accuracy for the infinite width predictions, \cite{neyshabur2018the} find that the performance of finite width CNNs improves as the width gets larger. 
These results appear in tension with our understanding of the scaling behavior of large width networks.
~
Second, there is prior empirical evidence suggesting the scaling behavior in CNNs is different from FC networks \cite{2019arXiv190206720L}.

To examine these differences between the observed behavior of CNNs and FCs, we present an extension of the formalism of \cite{dyer2020asymptotics} to convolutional networks. 
We study \emph{correlation functions} -- ensemble averages over weight configurations of quantities built out of the network map and its derivatives. 
We conjecture a simple scaling relation for correlation functions for models with convolutional, skip, dense, and global average pooling layers. 
We prove this relation for deep linear and and single hidden layer networks with smooth activations and check the relation empirically in a broader context including deep non-linear networks, ReLU networks, and networks with max pooling layers.
This scaling relation serves as the basis for bounding corrections to linear evolution and allows us to study how model performance depends on network width. 

\paragraph{Primary contributions}
\begin{itemize}
    \item We derive a set of scaling relations for correlation functions, a general class of expectation values of the network map and its derivative. These relations rely on generalizing previous diagramatic methods to convolutional networks. 
    
    \item We apply our scaling relations to the loss and accuracy during training. In particular we argue that the difference between full and linearized test loss scales as $\mathcal{O}(n^{-1})$.
    
    \item We confirm the predicted loss scaling empirically for deep non-linear networks trained on subsets of CIFAR-10 and find that it is consistent with finite width networks \emph{either} outperforming or underperforming their infinite width counterparts. 
    
    \item We further apply our relations to derive asymptotically tight bounds for the change of the NTK during training. 
    
\end{itemize}

\begin{figure*}[t!]
\centering
\captionsetup[subfigure]{oneside,margin={1.2cm,0cm}}

\subfloat[]{\label{fig:sa_intro}
    \includegraphics[scale=0.35]{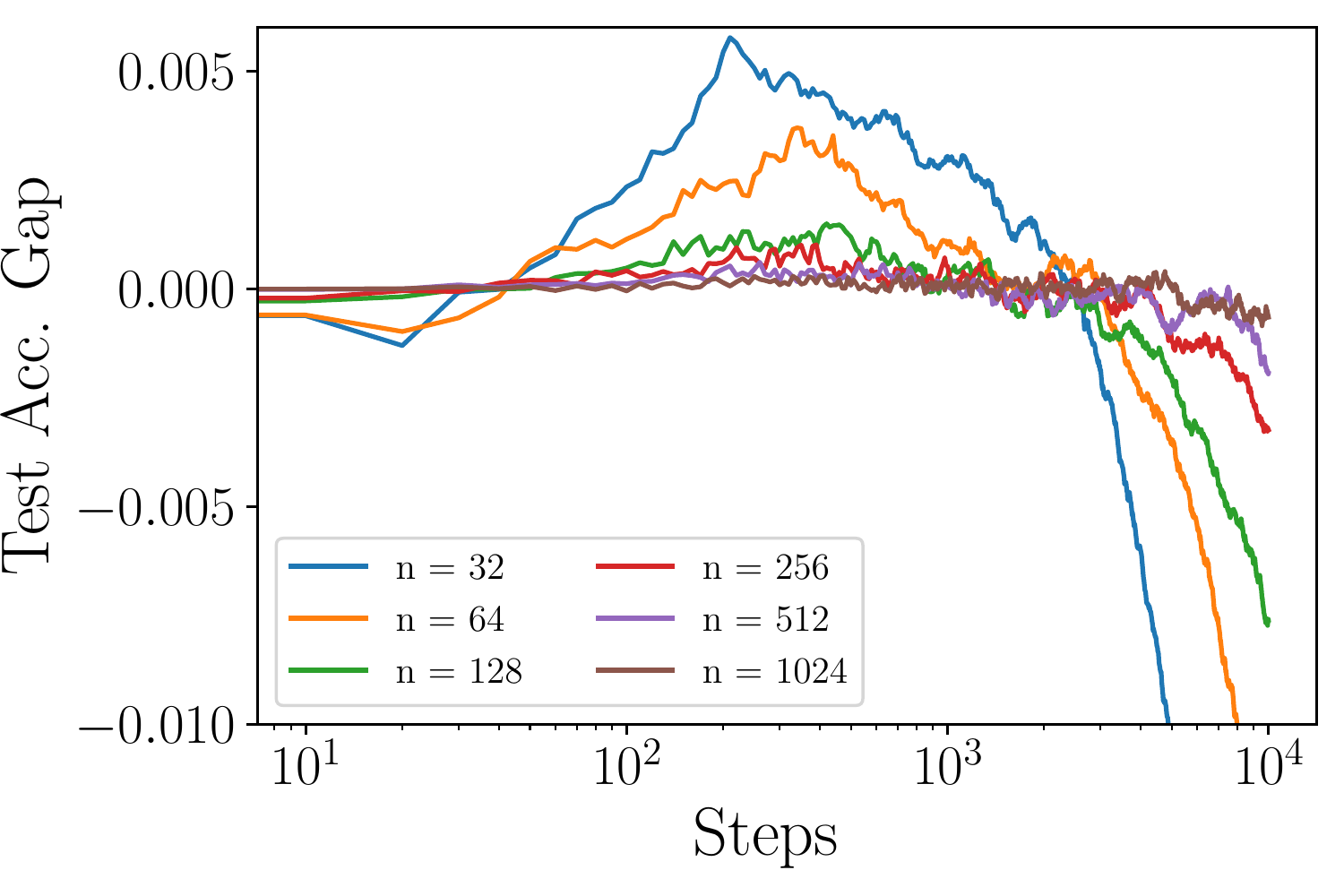}}
\subfloat[]{\label{fig:sd_intro}
    \includegraphics[scale=0.35]{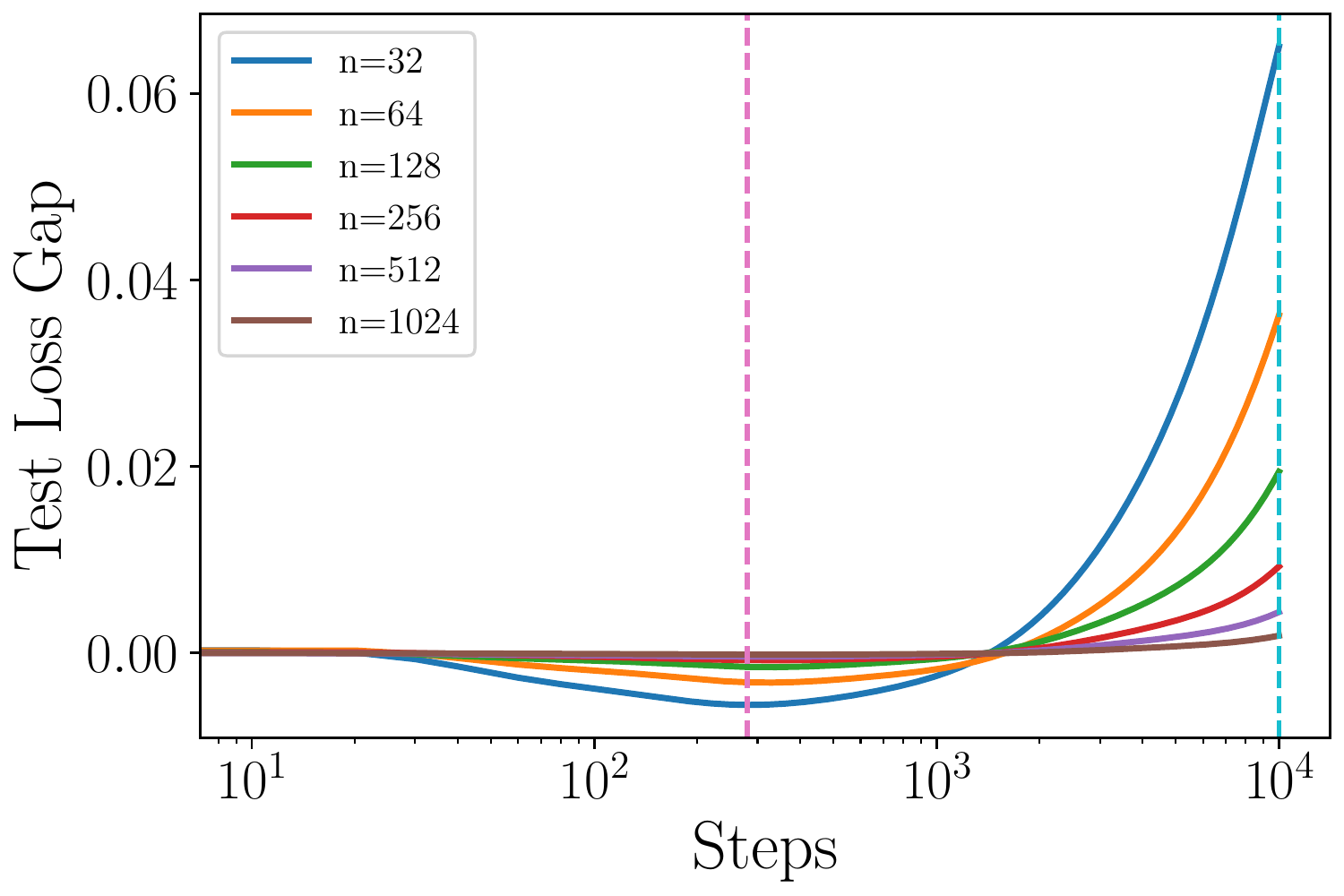}}
\subfloat[]{\label{fig:sc_intro}
    \includegraphics[scale=0.35]{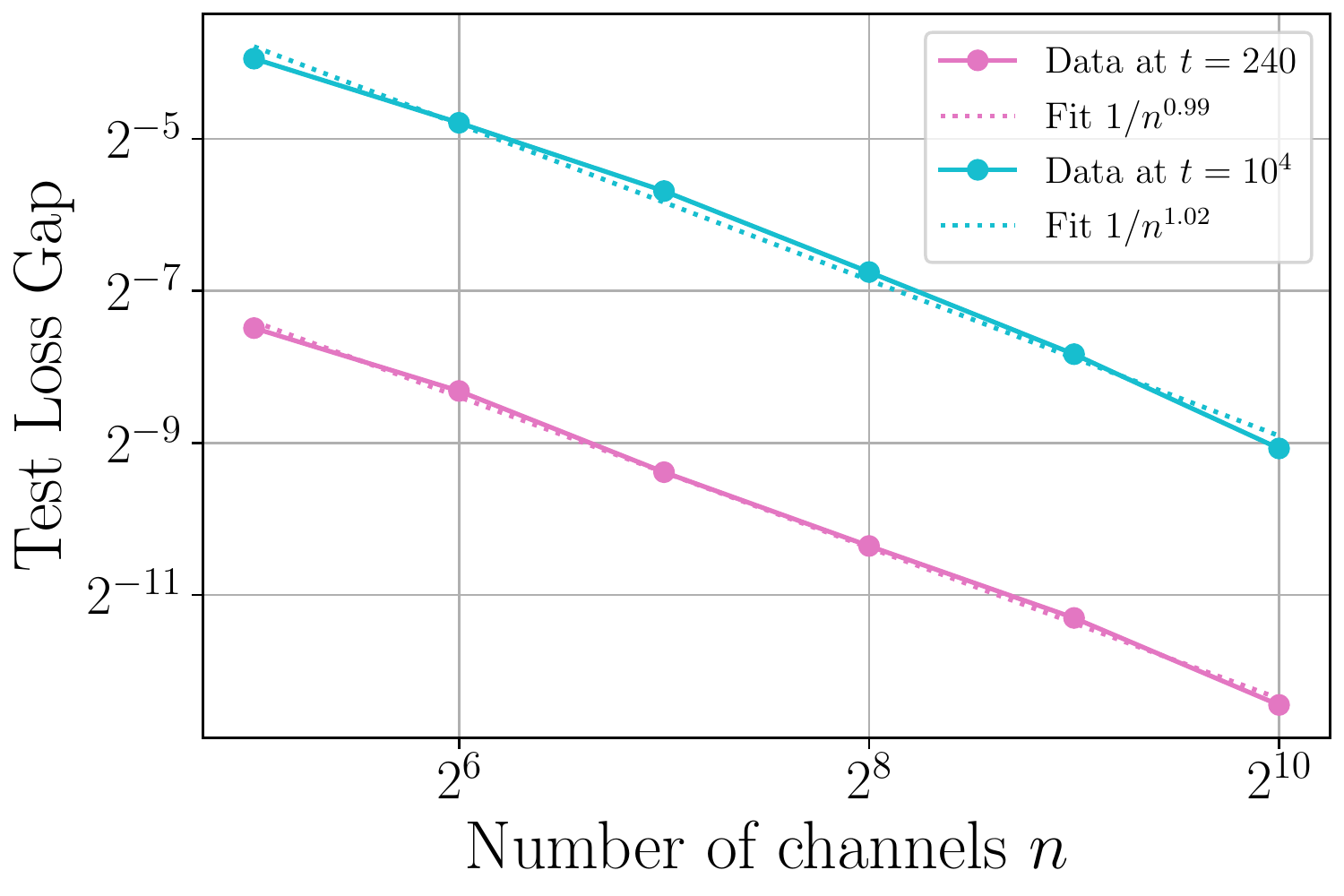}}
\caption{Difference in (a) accuracy and (b) loss between a finite width, four hidden layer CNN and the infinite width kernel prediction on a 2-class subset of CIFAR-10. The performance of finite width networks approach the infinite width performance as the network gets larger. The relative performance of infinite to finite width depends on the stopping criteria and finite width networks can either outperform or underperform their infinite width counterparts. (c) In either case, the difference between the finite width and infinite width losses scales as $O(n^{-1})$. 
}
\label{fig:loss_panel_intro}
\end{figure*}

\begin{figure*}[h!]
\centering
\subfloat{
  \includegraphics[width=0.47\textwidth]{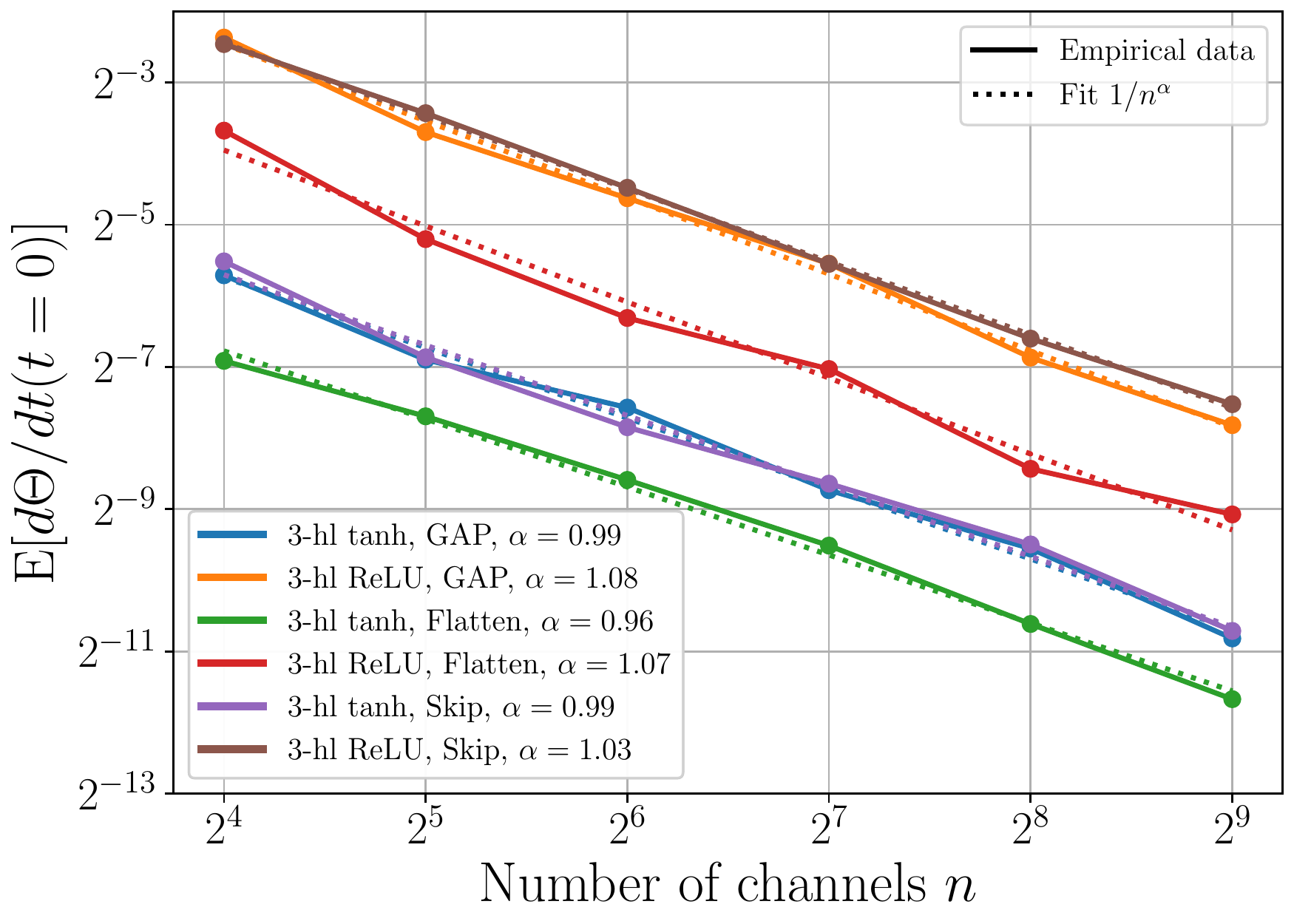}
  \label{fig:mean_dtheta_dt0_intro}
}
\subfloat{
  \raisebox{1.1in}{
  \begin{tabular}{c|c|c|c}
$\alpha$     & Flatten &   GAP  & Skip  \\\hline
1 hl $\tanh$ &  1.00 &  0.99  &  0.99 \\ 
1 hl ReLU    &  0.98 &  1.01  &  0.98 \\ 
3 hl $\tanh$ &  0.96 &  0.99  &  0.99 \\
3 hl ReLU    &  1.07 &  1.08  &  1.03 \\
\end{tabular}}
  \label{tab:mean_dtheta_dt0_intro}
}
\caption{
Example of empirical scaling relations matching the predicted $1/n$ scaling for $\lexpp{\theta}\frac{d\Theta_{0}}{dt}\rexp$ for a 3 hidden layer CNN with a flatten, global average pooling, or skip-connection operations and different activation functions. See \Sec{experiments} for details and more experiments.
}

\label{fig:basicscaling}

\end{figure*}

\section{Related work}
There has been significant progress understanding the behavior of wide neural networks. At initialization, wide networks of any depth behave as Gaussian processes \cite{Neal1996, lee2018deep, g.2018gaussian, novak2019bayesian, garriga-alonso2018deep, yang2019scaling}. Finite width corrections to the Gaussian process picture have been discussed in \cite{Yaida:2019sjo}.
The infinite width training of NTK parameterized networks was introduced in \cite{NTK} and studied in \cite{2019arXiv190206720L,du2018gradient,du2018gradient2,allen2018convergence,arora2019fine,daniely2017sgd}.
Wide convolutional networks have been studied in \cite{garriga-alonso2018deep,novak2019bayesian,xiao2018dynamical, arora2020harnessing,Li2019EnhancedCN,Arora2019OnEC}. \cite{yang2019tensor} study the infinite width limit of a more general class of network, that includes convolutional networks as a special case.
This work is most closely related to \cite{dyer2020asymptotics, Aitken2020OnTA} which also applied diagramatic techniques to derive asymptotic scaling rules for correlation functions. See also \cite{NIPS2017_6857} for another use of diagrams in this context. We also discuss finite width corrections to the evolution of convolutional networks, these corrections were studied for fully-connected networks in \cite{dyer2020asymptotics, huang2019dynamics}. Preliminary experiments for this work were run using the neural tangents python library 
\cite{neuraltangents2020}. During the completion of this work \cite{lee2020finite} appeared which performed extensive empirical studies comparing the performance of finite width CNNs and infinite width kernel methods.

\section{Theory}\label{sec:theory}
In this section we present our main theoretical result. Namely, a conjectured class of bounds, Conjecture~\ref{conj:main}, governing the scaling with respect to width of quantities built out of the network map and its derivatives. 

We consider neural networks, $f(x)$, built out of a collection of stacked activations with non-linearity $\sigma$,
\begin{align}
    &\alpha^{(\ell+1)}\,=\,\sigma\left(\tilde{\alpha}^{(\ell+1)}[\alpha^{(0)},\ldots,\alpha^{(\ell)}]\right)\,, \ \ \ \alpha^{(0)} =x\,,
\end{align}
where $\tilde{\alpha}^{(\ell+1)}$ is a pre-activation selected from one of the following layer types.
\begin{itemize}
\item \textbf{Convolutional layers}

For a convolutional layer with kernel size $k_{w}\times k_{h}$ and $n$ channels,
\begin{align}\label{eq:layer_conv}
\tilde{\alpha}^{(\ell+1)}_{r,s;i}&=\frac{1}{\sqrt{k_{w} k_{h} n}}\sum_{a=1}^{k_{w}}\sum_{b=1}^{k_{h}}\sum_{j=1}^{n}W^{(\ell)}_{a,b;ij} \alpha_{r+a,s+b;j}^{(\ell)}\,.
\end{align}
\item \textbf{Dense layers}
\es{eq:layer_dense}{
\tilde{\alpha}^{(\ell+1)}_{i}&=\frac{1}{\sqrt{\mathcal{W} \mathcal{H} n}}\sum_{r=1}^{\mathcal{W}}\sum_{s=1}^{\mathcal{H}}\sum_{j=1}^{n}W^{(\ell)}_{r,s;ij}\alpha^{(\ell)}_{r,s;j}\,.
}
In the case where the activation $\alpha^{(\ell)}_{i}$ has no spacial indices (i.e. after a dense or pooling layer, we drop the sum over $r$ and $s$ and the factor of $\mathcal{W}$, $\mathcal{H}$ in the normalization).
\item \textbf{Skip connections}

For skip connection connecting the pre-activation at layer $\ell+1$ to the activation of layer $\ell+1-k$ we have,
\es{eq:layer_skip}{
\tilde{\alpha}^{(\ell+1)}&=\tilde{\alpha}^{(\ell+1)}_{\textrm{straight}}+\alpha^{(\ell+1-k)}\,.
}
Here $\tilde{\alpha}^{(\ell+1)}_{\textrm{straight}}$ is the output of either a dense, \Eq{layer_dense},  or convolutional, \Eq{layer_conv}, operation. 
\item \textbf{Global Average Pooling (GAP)}
\es{gap_layer}{
\tilde{\alpha}^{(\ell+1)}_{i}&=\frac{1}{\mathcal{W} \mathcal{H}}\sum_{r=1}^{\mathcal{W}}\sum_{s=1}^{\mathcal{H}}\alpha_{r,s;i}^{(\ell)}\,.
}
\end{itemize}
We consider networks terminated by linear transform after a flatten or global average pooling operation,
\es{f_def}{
f_{\textrm{Flatten}}(x)&=\frac{1}{\sqrt{\mathcal{W} \mathcal{H} n}}\sum_{r=1}^{\mathcal{W}}\sum_{s=1}^{\mathcal{H}}\sum_{i=1}^{n}V_{r,s;i}\alpha^{(d)}_{r,s;i},\\
f_{\textrm{GAP}}(x)&=\frac{1}{\mathcal{W} \mathcal{H}\sqrt{n}}\sum_{r=1}^{\mathcal{W}}\sum_{s=1}^{\mathcal{H}}\sum_{i=1}^{n}V_{i}\alpha^{(d)}_{r,s;i}\,.
}

Following \cite{dyer2020asymptotics} we introduce a class of moments built out of the network map and its derivatives called correlation functions.
\begin{definition} \label{def:corr}
  A \emph{correlation function} , $C(x_{1},x_{2},\ldots,x_{m})$, is an ensemble average over products of the network map, $f(x)$, and its derivatives, $\frac{\partial^{k}f(x)}{\partial_{\mu_{1}}\partial_{\mu_{2}}\cdots\partial_{\mu_{k}}}$, subject to the condition that all derivatives are summed in pairs.
  A general correlation function $C$ takes the form
  \es{corr}{
    C:=&
    \!\! \sum_{\mu_1,\dots,\mu_{k_m}} \!\!
    \Delta_{\mu_1 \dots \mu_{k_m}}^{(\pi)}\\
    &\lexpp{\theta}
    \frac{\partial^{k_{1}}f(x_{1})}{\partial_{\mu_{1}}\cdots\partial_{\mu_{k_{1}}}}
    \frac{\partial^{k_{2}-k_{1}}f(x_{2})}{\partial_{\mu_{k_{1}+1}}\cdots\partial_{\mu_{k_{2}}}}
    \cdots
    \frac{\partial^{k_{m}-k_{m-1}}f(x_{m})}{\partial_{\mu_{k_{m-1}+1}}\cdots\partial_{\mu_{k_{m}}}}
    \rexp \,.
}
  Here, $0 \le k_1 \le \cdots \le k_{m-1} \le k_m$ are integers,\footnote{We adopt the convention that $k_a=k_{a-1}$ represents a factor of $f$ with no derivatives acting.} $m$ and $k_m$ are even, $\pi \in S_{k_m}$ is a permutation, and
  $
    \Delta_{\mu_1 \dots \mu_{k_m}}^{(\pi)} =
    \delta_{\mu_{\pi(1)} \mu_{\pi(2)}} \cdots \delta_{\mu_{\pi(k_m-1)} \mu_{\pi(k_m)}} \,.
  $
  We use $\delta$ to denote the Kronecker delta.
\end{definition}
We refer to paired summed indices as contracted derivatives and refer to factors of the network map with such contracted derivatives as being contracted in $C$.

Some examples include.
\es{examples}{
\lexpp{\theta}f(x_1)f(x_2)\rexp,\, \lexpp{\theta}\sum_{\mu}\frac{\partial f(x_1)}{\partial\theta_{\mu}}\frac{\partial f(x_2)}{\partial\theta_{\mu}}\rexp\,,\\
\lexpp{\theta}\sum_{\mu,\nu}\frac{\partial^{2} f(x_1)}{\partial\theta_{\mu}\partial\theta_{\nu}}\frac{\partial f(x_2)}{\partial\theta_{\mu}}\frac{\partial f(x_3)}{\partial\theta_{\nu}}f(x_4)\rexp
}
For every such correlation function we define the associated cluster graph.
{\definition \label{def:clust} The cluster graph, $G_{C}(V,E)$, associated to a correlation function $C(x_{1},x_{2},\ldots,x_{m})$, is a graph consisting of $m$ vertices, one corresponding to each factor of the network map in Equation~\eqref{corr}. The graph has a single edge between vertices corresponding to factors of the network map sharing a pair of contracted derivatives.
\es{graph_def}{
V&=\{v_{1},v_{2},\ldots,v_{m}\}\\
E&=\{(v_{i},v_{j}): \textrm{For $f(x_{i})$ and $f(x_{j})$ contracted in $C$}\}\,.
}
}
In \cite{dyer2020asymptotics} the authors argue for a set of bounds on correlation functions for fully connected networks based on the cluster graph. For a correlation function $C$ with a cluster graph containing $n_{e}$ even components (connected components with an even number of vertices), $n_{o}$ odd components, and $m$ vertices, they argue that the correlation function satisfies $C=\mathcal{O}(n^{n_e+n_o/2-m/2})$. The authors prove these bounds for deep linear and one-hidden non-linear fully connected networks and check empirically in a variety of contexts. Subsequently \cite{Aitken2020OnTA} established this bound for deep non-linear networks with polynomial activations. Here we extend this to networks with convolution, skip, and global average pooling layers. We present a proof in the deep-linear and 1-hidden layer non-linear case. 
\begin{conjecture}\label{conj:main}
  Let $C(x_1,\dots,x_m)$ be a correlation function with cluster graph, $G_C$.
  Suppose that $G_C$ has $n_e$ connected components with an even size, and $n_o$ components of odd size,
  then $C(x_1,\dots,x_m) = \mathcal{O}(n^{s_C})$, where
  \begin{align}
    s_C = n_e + \frac{n_o}{2} - \frac{m}{2} \,. \label{eq:s}
  \end{align}
\end{conjecture}
We have tested this conjecture empirically in a variety of contexts, and these results appear in Section~\ref{sec:experiments}. We are also able to prove Conjecture~\ref{conj:main} for deep linear and one-hidden-layer non-linear networks.
\begin{thm}\label{thm:main}
  Conjecture~\ref{conj:main} holds for deep linear and one-hidden-layer networks with smooth activations.
\end{thm}
The argument for the deep linear case is summarized below. A detailed proof for deep linear networks and one-hidden-layer non-linear networks appears in the Supplement.

In the deep linear case, the proof follows from two lemmas.
\begin{lemma}\label{lemma:decomp}
Let $f(x)$ be a deep linear network of depth $d$ made up of convolutional, skip, dense, and GAP layers. Then the network function can be written as a finite sum over $\mathcal{N}_{f}$ functions $f_{I}(x)$, where each function $f_{I}$ has the topology of a fully connected network with depth $d_{I}\leq d$.
\es{network_decomp}{
f(x)&=\sum_{I=1}^{\mathcal{N}_{f}}f_{I}(x)\,.
} 
Furthermore, let $\{\theta_{I}\}$ be the set of weights of each $f_{I}(x)$ and $\{\theta\}$ the weights of $f$. Then, $\{\theta_{I}\}\in\{\theta\}$.
\end{lemma}

The decomposition of the network map motivates a generalization of the definition of correlation functions to expectations involving the maps $f_{I}(x)$. We dub such expectation values \emph{mixed correlation functions}.
\begin{definition} \label{def:corr_mixed}
  A \emph{mixed correlation function}, $C_{I_{1},I_{2},\ldots,I_{m}}(x_{1},x_{2},\ldots,x_{m})$, is an ensemble average over products of the functions, $f_{I}(x)$, and its derivatives, $\frac{\partial^{k}f_{I}(x)}{\partial_{\mu_{1}}\partial_{\mu_{2}}\cdots\partial_{\mu_{k}}}$, subject to the condition that all derivatives are summed in pairs.
\end{definition}

The decomposition of the network map in Lemma~\ref{lemma:decomp} allows us to write correlation functions of CNNs in terms of finite sums over correlation functions of fully connected networks and thus reuse much of the technology introduced in \cite{dyer2020asymptotics}.  This is formalized in the following lemma.
\begin{lemma}\label{lemma:mixed_bound}
Let $C_{I_{1},I_{2},\ldots,I_{m}}(x_{1},x_{2},\ldots,x_{m})$ be a mixed correlation function. Let $G_{C}$ be a cluster graph associated to $C_{I_{1},I_{2},\ldots,I_{m}}(x_{1},x_{2},\ldots,x_{m})$ via Definition~\ref{def:clust} ignoring the labels $\{I_{1},I_{2},\ldots,I_{m}\}$. Let $n_{e}$, $n_{o}$ be the number of even, odd clusters in $G_{C}$, then $C_{I_{1},I_{2},\ldots,I_{m}}(x_{1},x_{2},\ldots,x_{m})=\mathcal{O}(n^{s_C})$ with
\begin{align}
    s_{C}=n_e+\frac{n_o}{2}-\frac{m}{2}\,.
\end{align}
\end{lemma}

Together Lemma~\ref{lemma:decomp} and Lemma~\ref{lemma:mixed_bound} imply Theorem~\ref{thm:main} for deep linear networks. Lemma~\ref{lemma:decomp} follows from the definitions of our layers above. Here we show this for convolution layers, and leave the details of skip and GAP layers to the Supplement.
\begin{proof}
(Lemma~\ref{lemma:decomp} -- convolution layers). Consider a network with a convolution layer at layer $\ell+1$. We write the network as $f(x) = g(\tilde{\alpha}^{(\ell+1)}_{r,s;i})$, where $g$ is the map from layer $\ell+1$ to the output.
\begin{align}\label{eq:conv_decomp}
    f(x) &= g(\tilde{\alpha}^{(\ell+1)}_{r,s;i})\,=\, \sum_{a=1}^{k_{w}}\sum_{b=1}^{k_{h}}g(\frac{1}{\sqrt{n}}\sum_{j=1}^{n}W_{a,b;ij} \alpha_{r+a,s+b;j}^{(\ell)})\nonumber\\
    &=\sum_{I=1}^{k_{w}\times k_{h}} f_{I}(x)\,, \ \ \ I=\{a,b\}.
\end{align}
In the first line, we used the fact that $f$ is linear to move the sum over the kernel outside of $g$. In the last line we let $I$ run over all terms in the double sum over the filter. We have dropped the $1/\sqrt{k_{w}k_{h}}$ normalization factor for clarity, as it does not effect the argument. Each $f_{I}(x)$ is a network with the convolutional layer at depth $\ell+1$ replaced with a dense layer. Lemma~\ref{lemma:decomp} follows from repeating this expansion for all convolutions.
\end{proof}

We prove Lemma~\ref{lemma:mixed_bound} in the Supplement. We rely on the graphical Feynman diagram techniques introduced in \cite{dyer2020asymptotics}.

\subsection{Evolution}
Conjecture~\ref{conj:main} has important implications for the training dynamics of convolutional networks at both infinite and finite width.

Consider a network trained via gradient flow with mean squared error (MSE) loss.
\begin{align}
    \frac{d\theta_{\mu}}{dt}=-\frac{\partial L_{\textrm{train}}}{\partial\theta_{\mu}}\,, \ \ \ L_{\textrm{train}}=\frac{1}{2}\sum_{a\in\mathcal{D}_{\textrm{train}}}\left(f(x_a)-y_a\right)^{2}\,.
\end{align}
In function space, we have,
\begin{align}
  \frac{df(x;t)}{dt}=-\sum_{x_{a}\in\mathcal{D}_{\textrm{train}}}\Theta(x,x_{a};t)\left(f(x_{a};t)-y_{a}\right)\,.
\label{eq:dfdt_full}
\end{align}
In general this equation can describe quite complicated training dynamics, as a result of the time dependence of $\Theta(t)$. Empirically we find some non-trivial late time behavior for certain CNN models which has to be treated with care. We elaborate on this in the Supplement. At infinite width, however, $\Theta$ is constant and the dynamics reduce to that of training a linear model. 

As we will explain, Conjecture~\ref{conj:main} bounds the change in the kernel as
\begin{align}\label{eq:theta_change}
    \lexpp{\theta}\Theta(t)-\Theta(0)\rexp=\mathcal{O}(n^{-1})
\end{align}
 In Figure~\ref{fig:3hl_theta_fit} and \Tab{fit_vals_theta_evol} we see evidence that this bound is saturated in a variety of CNNs.

\begin{figure}
    \centering
    \includegraphics[width=0.5\textwidth]{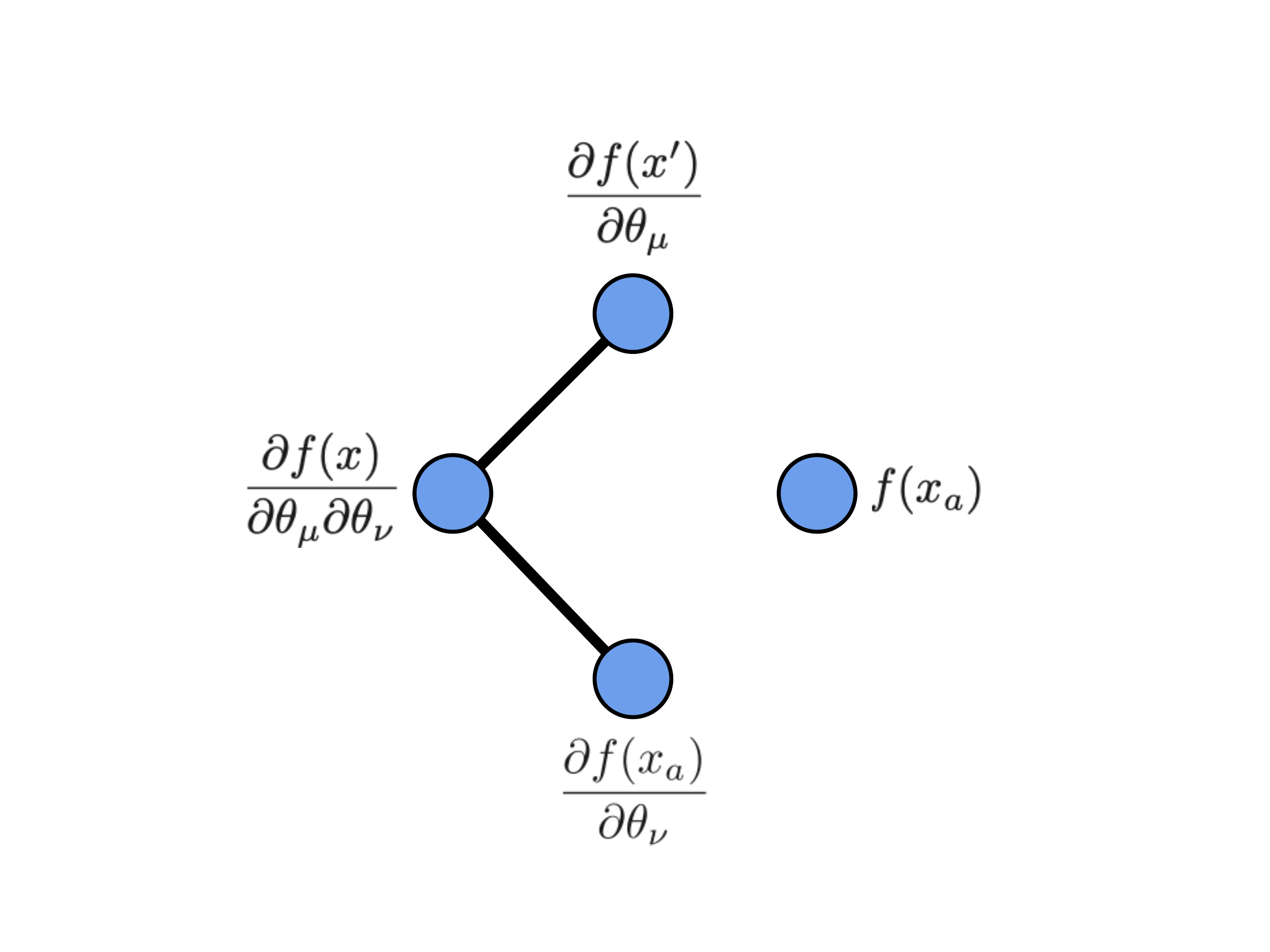}
    \caption{Cluster graph for $C=\lexpp{\theta}\frac{\partial^2 f(x)}{\partial\theta_{\mu}\partial\theta_{\nu}}\frac{\partial f(x')}{\partial\theta_{\mu}}\frac{\partial f(x_{a})}{\partial\theta_{\nu}}f(x_{a})\rexp$}
    \label{fig:clusterGraph}
\end{figure}

To understand this analytically, we begin with an illustrative example. Consider the time derivative of the NTK,
\begin{align}
  \lexpp{\theta}\frac{d\Theta}{dt}\rexp &=-\sum_{\mu,\nu,a}\lexpp{\theta}\frac{\partial^2 f(x)}{\partial\theta_{\mu}\partial\theta_{\nu}}\frac{\partial f(x')}{\partial\theta_{\mu}}\frac{\partial f(x_{a})}{\partial\theta_{\nu}}f(x_{a})\rexp\\
  & \ + x\leftrightarrow x'\nonumber.
\end{align}
Here the notation $x\leftrightarrow x'$ indicates symmetrization with respect to the arguments $x$ and $x'$. 
This expression is a sum of correlation functions, $C(x,x',x_{a},x_{a})=\lexpp{\theta}\frac{\partial^2 f(x)}{\partial\theta_{\mu}\partial\theta_{\nu}}\frac{\partial f(x')}{\partial\theta_{\mu}}\frac{\partial f(x_{a})}{\partial\theta_{\nu}}f(x_{a})\rexp$. The cluster graph corresponding to each $C$ is shown in Figure~\ref{fig:clusterGraph}. The graph has four vertices and two odd clusters. Conjecture~\ref{conj:main} then gives $C=\mathcal{O}(n^{-1})$.

In \cite{dyer2020asymptotics} it was shown that the cluster graph for the correlation functions corresponding to expectations of all higher order time derivatives $\lexpp{\theta}\frac{d^{k}\Theta}{dt^{k}}\rexp$ also satisfy $s_{C}\leq-1$. Combining this with Conjecture~\ref{conj:main} and assuming analyticity of $\Theta(t)$ gives \Eq{theta_change}.
\paragraph{Beyond infinite width}
The constancy of the NTK is a striking feature of infinite width networks. However in practice we mostly consider finite width networks and the connection between infinite and finite width evolution is not immediately clear. In \cite{dyer2020asymptotics, huang2019dynamics} the authors take steps towards understanding finite width networks. In particular they show that the scaling relations, Conjecture~\ref{conj:main}, imply a systematic expansion for the evolution of the network map.
\begin{align}\label{eq:pert}
    f(t;x)=f^{(0)}(x;t)+\frac{1}{n}f^{(1)}(x;t)+\cdots\,,
\end{align}
Here, $f^{(0)}(x;t)$ is the linearized infinite width evolution, Equation~\eqref{linearevol}, 
and the higher order terms can be iteratively solved for in terms of the network map at initialization. 
The derivation of this result relies only on Conjecture~\ref{conj:main} and so applies here as well. Some important consequences of this expansion are scaling relations for the loss and accuracy during training, which we now describe.

\begin{figure*}[h!]
\centering
\subfloat[$\lexpp{\theta}\frac{d\Theta_{0}}{dt}\rexp$]{
  \includegraphics[width=0.47\textwidth]{figs/dTheta_dt_0_3hl.pdf}
  \label{fig:mean_dtheta_dt0}
}
\subfloat[$\lvar{\theta}\Theta_{0}\rexp$]{
  \includegraphics[width=0.47\textwidth]{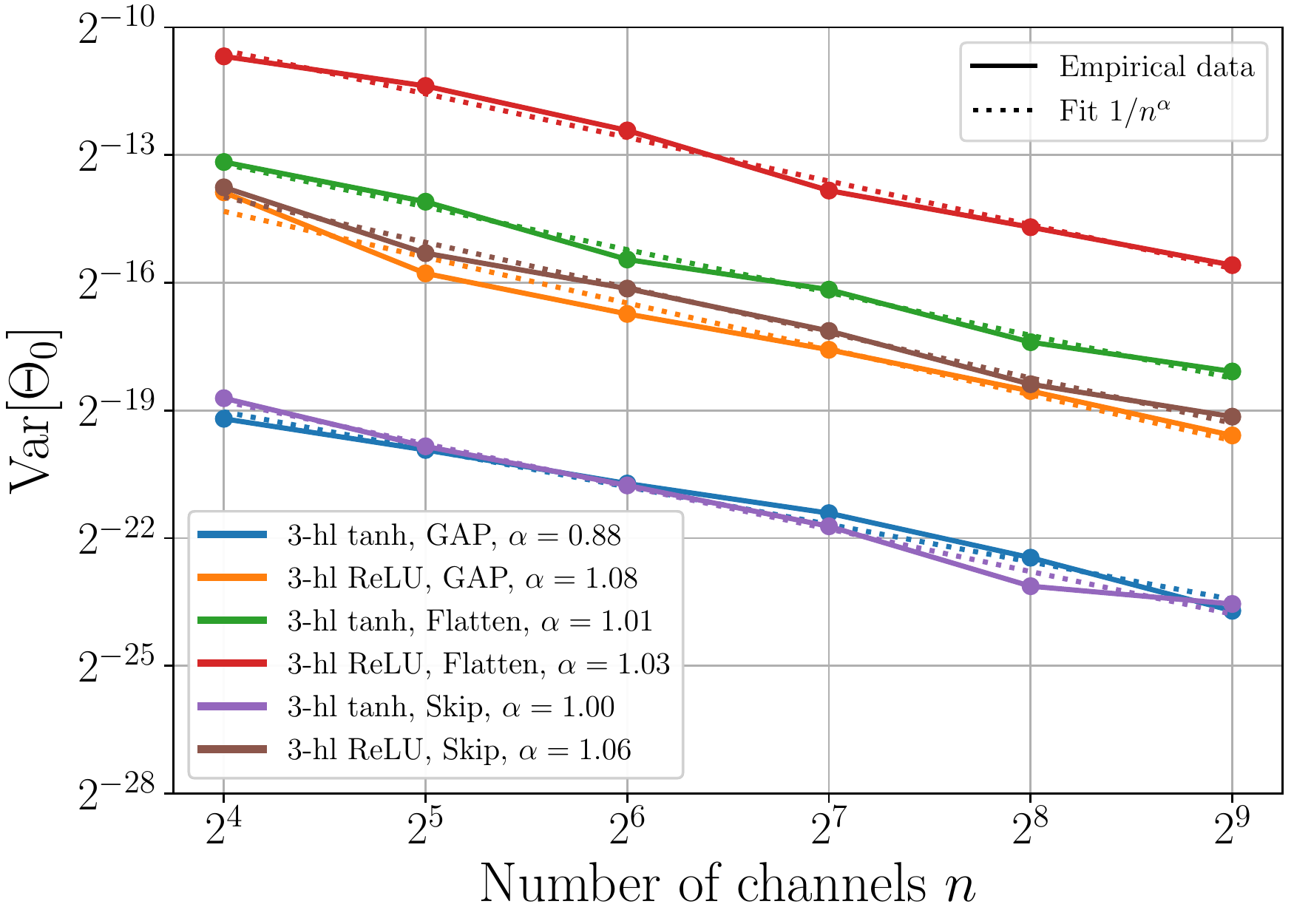}
  \label{fig:main_var_theta0}
}

\subfloat[$\lexpp{\theta}\frac{d\Theta_{0}}{dt}\rexp$]{
  \begin{tabular}{c|c|c|c}
$\alpha$     & Flatten &   GAP  & Skip  \\\hline
1 hl $\tanh$ &  1.00 &  0.99  &  0.99 \\ 
1 hl ReLU    &  0.98 &  1.01  &  0.98 \\ 
3 hl $\tanh$ &  0.96 &  0.99  &  0.99 \\
3 hl ReLU    &  1.07 &  1.08  &  1.03 \\
\end{tabular}
  \label{tab:mean_dtheta_dt0}
}
\subfloat[$\lvar{\theta}\Theta_{0}\rexp$]{
   \begin{tabular}{c|c|c|c}
$\alpha$     & Flatten &   GAP  & Skip  \\\hline
1 hl $\tanh$ &  1.07 &  1.03  & 1.06    \\ 
1 hl ReLU    &  1.02 &  0.98  & 1.02    \\ 
3 hl $\tanh$ &  1.01 &  0.88  & 1.00   \\
3 hl ReLU    &  1.03 &  1.08  & 1.07   \\
\end{tabular}
  \label{tab:var_theta0}
}

\caption{Scaling of (a) $\lexpp{\theta}d\Theta_{0}/dt (t=0)\rexp$ and (b) $\lvar{\theta}\Theta_{0}\rexp$ with number of channels for three-hidden-layer networks with a few choices of layer types and activation function. Shown is the empirical data and fit $1/n^\alpha$ where $n$ is the number of channels. Tables (c) and (d) shows additional fitted values for one-hidden-layer networks. Models with one hidden layer was fit in the range $n=[16, 32, ..., 2048]$, and models with three hidden layers in the range $n=[16,32, ..., 512]$. The mean and variance are calculated over 100 initializations.}

\label{fig:basicscaling}

\end{figure*}

\subsection{Performance scaling}\label{sec:theory_loss}
One natural application of the large width evolution in \Eq{pert} is understanding the dynamics of the loss and accuracy during training of wide networks. 
This question is at the heart of the generalization paradox, the observation that over-parameterized networks suffer no degradation in performance as they become larger \cite{neyshabur2018the}. It also describes the asymptotic behavior of the so called double descent curve \cite{Neyshabur2014InSO, Advani2017HighdimensionalDO,geiger2019scaling,Belkin15849,Geiger2019TheJT,Nakkiran2019DeepDD}.

The expansion of the network map in powers of $1/n$ leads to a corresponding expansion in the test loss
\begin{align}
    L_{\textrm{test}}(f)=&L_{\textrm{test}}(f^{(0)})+\frac{1}{n}\sum_{a\in\mathcal{D}_\textrm{test}}(f^{(0)}(x_{a})-y_{a})f^{(1)}(x_{a})\nonumber\\
    & \ \ \ +\mathcal{O}(n^{-2})\,. \label{eq:loss_expansion}
\end{align}
Thus we expect the difference between the full model and linearized test loss to scale as
\begin{align}\label{eq:loss_diff}
\lexpp{\theta}|L_{\textrm{test}}-L_{\textrm{test}}^{\textrm{lin}}|\rexp=\mathcal{O}(n^{-1})    
\end{align}
Here $L_{\textrm{test}}^{\textrm{lin}}:=L_{\textrm{test}}(f^{(0)})$. This scaling has been observed empirically in fully connected networks \cite{geiger2019scaling} and is the same scaling predicted in \cite{NIPS2008_3495, Advani2017HighdimensionalDO, 2019arXiv190805355M} for linear models. Here we see good agreement with this asymptotic behavior of the loss in deep convolutional networks (see Figure~\ref{fig:loss_panel}). 

This relation between the loss of a finite width network and the infinite width loss is of particular interest for CNNs. As mentioned above, convolutional networks often exhibit a gap in performance between infinite width networks and their finite width counterparts \cite{Arora2019OnEC, novak2019bayesian}. It is natural to ask whether we can understand this performance gap within the framework of the perturbative expansion around large width.

The relation, \Eq{loss_diff}, implies that the full and linearized loss approach each other at infinite width, however the relative ordering is not dictated. We will see in the models studied below, that either ordering is possible depending on the time during training. In particular, in the setup studied here, finite width models outperform their linear counterparts if training is stopped at the non-linear early stopping time. We expand further on the dynamics and performance of the particular models studied in \Sec{exp_loss}.

\section{Numerical Experiments}\label{sec:experiments}
To support the theoretical predictions in \Sec{theory}, we present numerical results for the scaling of the NTK at initialization as well as at convergence (100\% training accuracy) for one- and three-hidden-layer convolutional neural networks with layers of the types defined in \Sec{theory}. 
All models are trained on 2-class MNIST (0's and 1's) with 10 examples per class, with the exception of \Sec{exp_loss} which is trained on 2-class CIFAR (airplane and automobile) with 100 examples per class and tested on the full 2-class test dataset. Training was done with full-batch gradient descent and all models achieved 100\% training accuracy. The learning rate used (unless otherwise specified) was $0.25\cdot \frac{1}{\max{\lambda_i}}$, where $\lambda_i$ are the eigenvalues of the NTK.  

\subsection{Asymptotic scaling at initialization}
First, we consider the scaling of the correlation function corresponding to the time derivative of the NTK at initialization,
$C = \lexpp{\theta}\frac{d\Theta}{dt}\rexp$. As discussed above we expect this correlation function, and indeed all higher time derivatives to be $\mathcal{O}(n^{-1})$. In \Fig{basicscaling},  sub-panel~\ref{fig:mean_dtheta_dt0} gives empirical evidence of this for a variety of three-hidden-layer networks and further examples are listed in sub-panel~\ref{tab:mean_dtheta_dt0}.

Next we consider the expectation of the variance of the NTK. This variance can be written as the difference of two correlation functions,
\begin{align}
    \lvar{\theta}\Theta(x,x') \rexp  = \lexpp{\theta}\Theta(x,x')^{2}\rexp-\lexpp{\theta}\Theta(x,x')\rexp^{2}\,.
\end{align}
From Conjecture~\ref{conj:main}, each of these correlation functions is $\mathcal{O}(n^{0})$, thus we can bound $\lvar{\theta}\Theta \rexp=\mathcal{O}(n^{0})$.
In the Supplement we show that for deep linear networks and for one-hidden-layer networks, we can actually do better giving $\lvar{\theta}\Theta \rexp=\mathcal{O}(n^{-1})$.
In the spirit of Conjecture~\ref{conj:main}, we predict this scaling more generally. Note that the suppression of the variance of the NTK with width is crucial for $\Theta$ to have a well defined infinite width limit.
In particular, this implies typical realizations of the kernel will be close to the mean as the width increases. Sub-panel~\ref{fig:main_var_theta0} of \Fig{basicscaling} corroborates this predicted scaling for three-hidden-layer CNNs with more examples in sub-panel~\ref{tab:var_theta0}.

\subsection{Change in the NTK}
Next, we consider how the deviation in the NTK from initializaion depends on the number of color channels. Equation \eqref{eq:theta_change} predicts that this difference scales as $\mathcal{O}(n^{-1})$.   
As discussed above, this constancy of the kernel is what underlies the large width linear dynamics.

\begin{figure}[h!]
    \centering
    \includegraphics[width=0.5\textwidth]{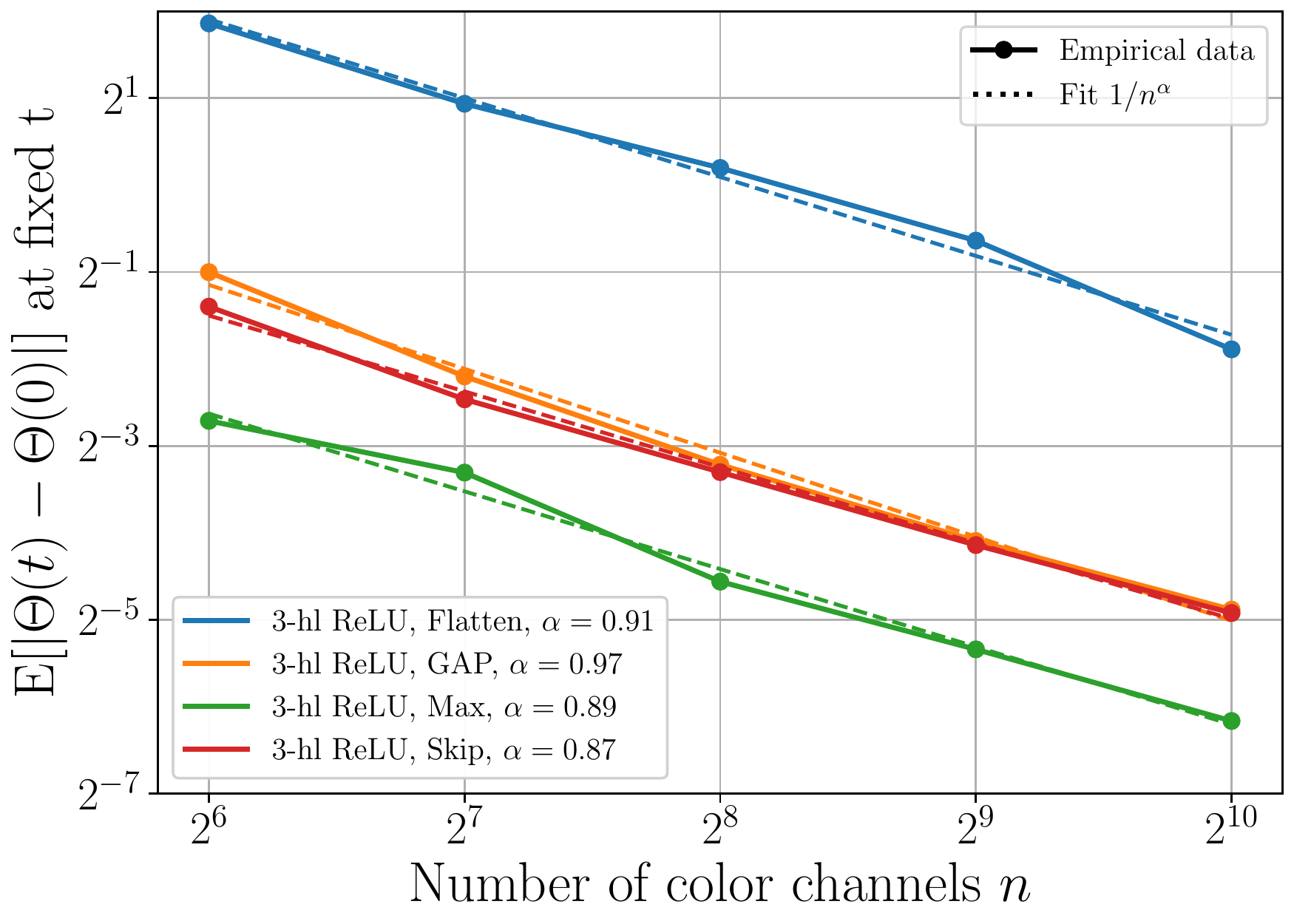}
    \caption{Scaling with number of channels $n$ for three-hidden-layer CNNs with ReLU activations. The scaling with $n$ is measured, for each choice of architecture, when all of the 10 random initializations have reached 100\% training accuracy.  }
    \label{fig:3hl_theta_fit}
\end{figure}

\Fig{3hl_theta_fit} shows that the scaling of the expected deviation from initialization of the NTK for a three-hidden-layer convolutional networks for different layer types and ReLU activation function. 
For each architecture, the fit is done at the time step when all the models for the 10 initializations have hit 100\% training accuracy. Further results are recorded in Table~\ref{tab:fit_vals_theta_evol}.

\begin{figure}[h!]
    \centering
    \includegraphics[width=0.5\textwidth]{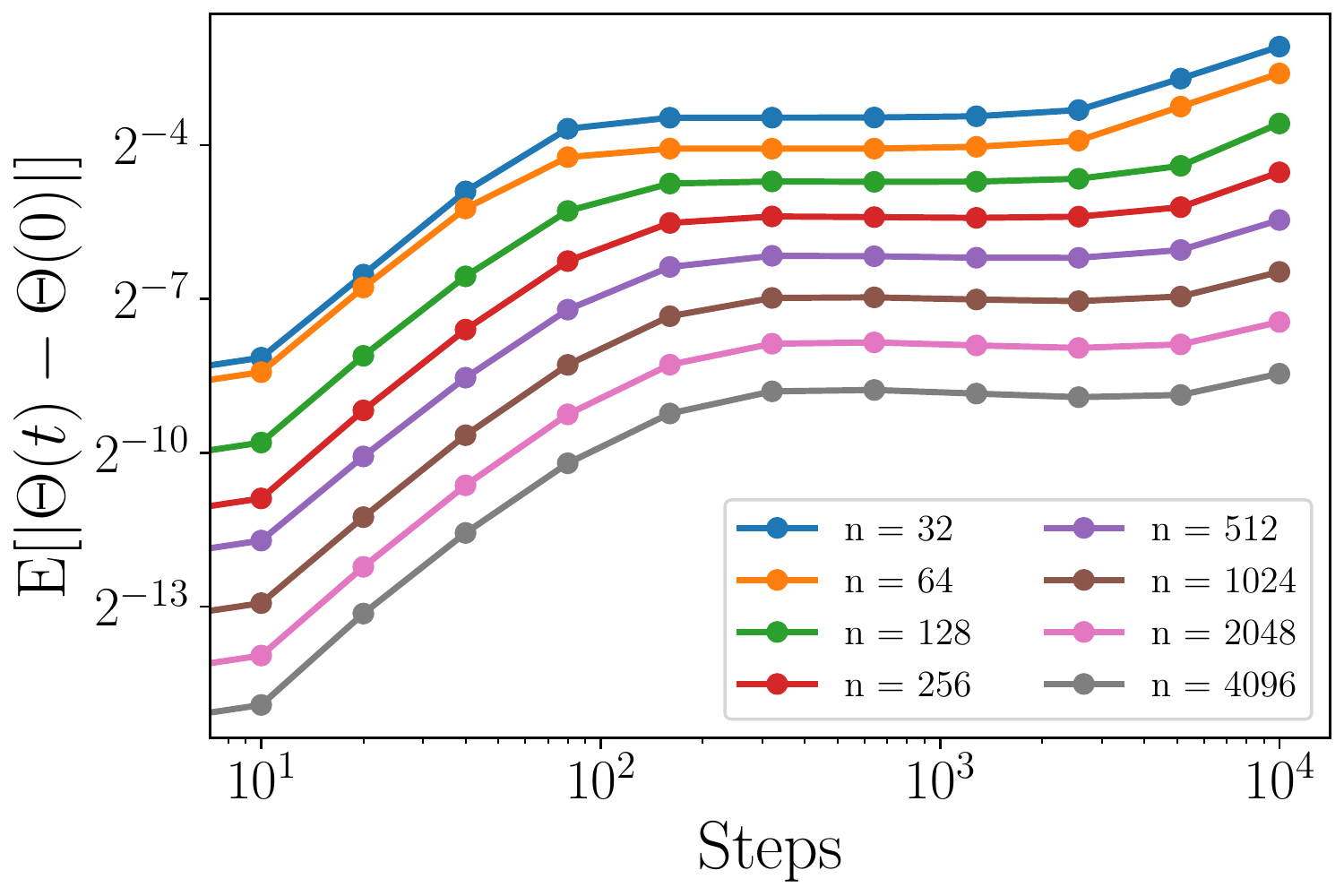}
    \caption{Evolution of $\lexpp{\theta}\Theta(t)-\Theta(0)\rexp$ for a convolutional network with one hidden layer with ReLU activation function and global average pooling. Expectation value is calculated over 10 random initializations.
    }
    \label{fig:theta_evolution}
\end{figure}

\begin{SCfigure}
    \centering
    \includegraphics[width=0.5\textwidth]{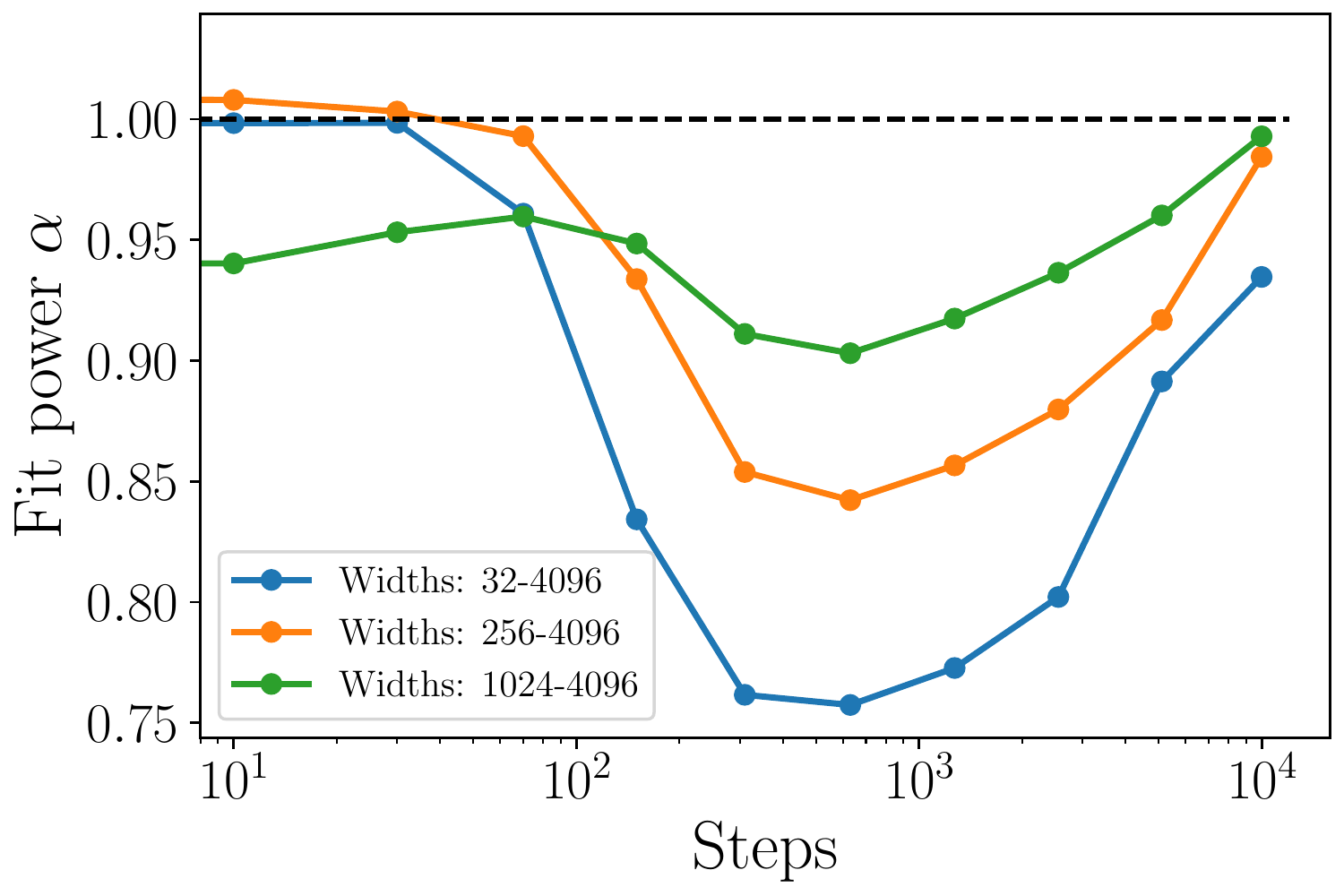}
    \caption{Evolution of the fit coefficient as a function of training steps in \Fig{theta_evolution}. Only considering larger values of $n$ gets result closer to the theoretical $1/n$ prediction as higher order terms become smaller. }
    \label{fig:fitpower_vs_time}
\end{SCfigure}

\Fig{theta_evolution} shows the evolution of $\Theta(t)$ during training with gradient descent for a single hidden layer convolutional network with ReLU activation function and global average pooling over a range of widths, $n$. 
\Eq{theta_change} is an asymptotic statement.
In practice, to see this scaling it is often necessary to go to $n\gg 32$.
As an example of the sensitivity to width, we look at the $1/n^\alpha$ fit over large and small ranges of $n$ in Figure~\ref{fig:fitpower_vs_time} during training.
We find greater deviations from $\mathcal{O}(n^{-1})$ scaling when including smaller widths. More generally, experimentally testing convolutional networks with enough channels to convincingly confirm or rule out Conjecture~\ref{conj:main} can be a challenge and we detail this in the Supplement.

\setcounter{table}{2}
\begin{table}[h!]
\centering
\begin{tabular}{c|c|c|c|c}
$\alpha$     & Flatten &   GAP  & Skip  & Max  \\ \hline
1 hl $\tanh$ &  0.87 &  0.86  &  0.89 & 0.88 \\ 
1 hl ReLU    &  0.92 &  0.88  &  0.84 & 0.71 \\ 
3 hl $\tanh$ &  1.02 &  1.15  &  1.14 & 0.81 \\
3 hl ReLU    &  0.91 &  0.97  &  0.87 & 0.89 \\
\end{tabular}
\caption{Fit power $\alpha$ in $1/n^\alpha$ for the number of channels $n$ for E$[|\Theta(t)-\Theta(0)|]$ for one- and three-hidden-layer CNNs with different layer types and activation functions. The one(three)-hidden-layer models are fitted in the range $n=[512, 1024, 2048,40962]$ ($n=[16, 32, .., 512]$). Expectation value is calculated over 10 random initializations.}
\label{tab:fit_vals_theta_evol}
\end{table}

\newpage
\subsection{Loss scaling}\label{sec:exp_loss}
In this section we empirically study the performance scaling discussed in \Sec{theory_loss} for a four-hidden-layer CNN trained on 2-class CIFAR with 100 examples per class. These models exhibit the now familiar property that their performance does not get worse as the number of channels is increased.

\begin{figure*}[th!]
\centering
\captionsetup[subfigure]{oneside,margin={1.2cm,0cm}}
\subfloat[]{\label{fig:sb}
    \includegraphics[scale=0.38]{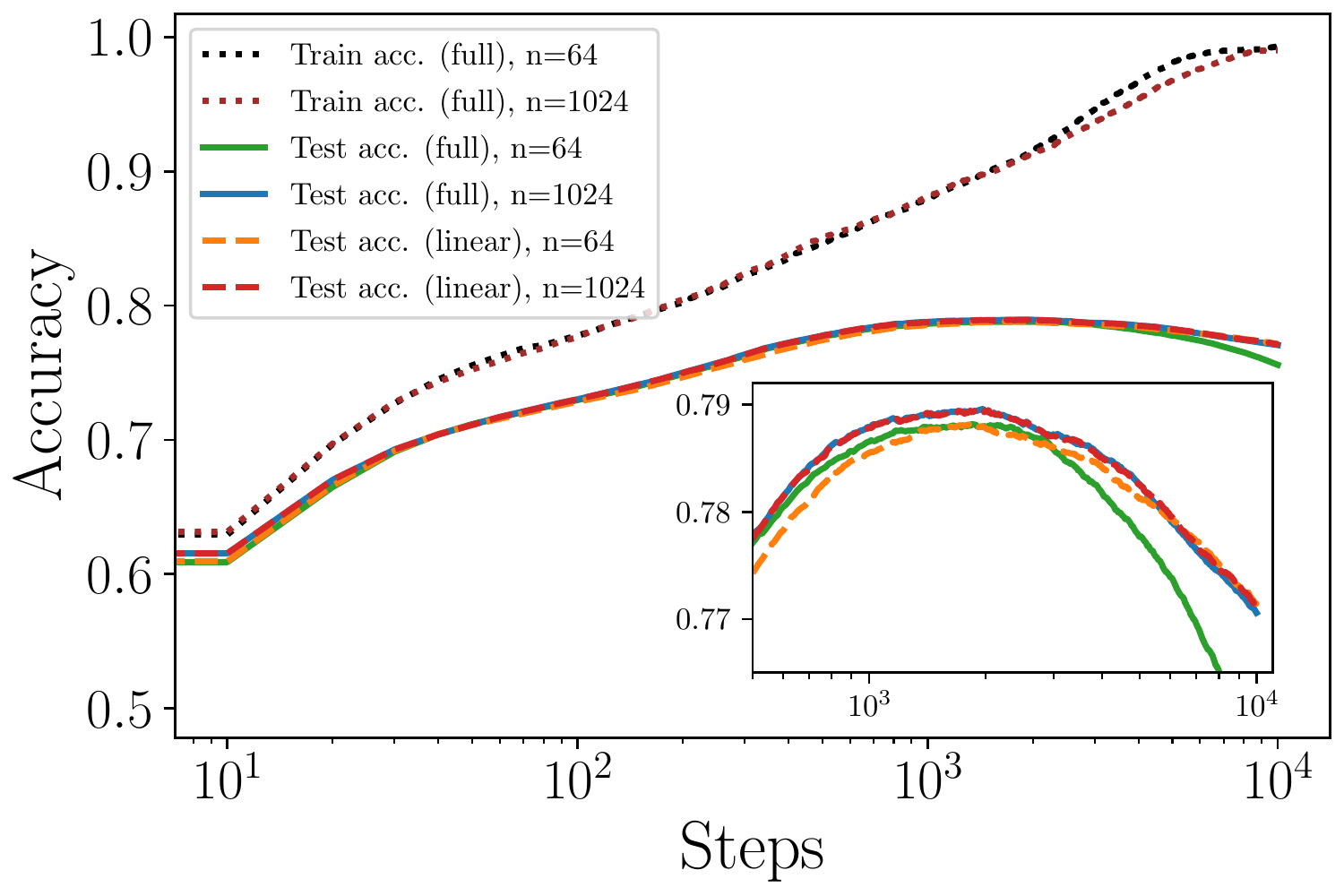}}
\subfloat[]{\label{fig:sa}
    \includegraphics[scale=0.38]{figs/TestAccGap.pdf}}

\subfloat[]{\label{fig:sc}
    \includegraphics[scale=0.38]{figs/TestLossFit.pdf}}
\subfloat[]{\label{fig:sd}
    \includegraphics[scale=0.38]{figs/TestLossGap.pdf}}
\caption{
Loss and accuracy curves for a convolutional network with four hidden layers with tanh activation functions and a flatten before the final dense layer trained on 2-class CIFAR with 100 examples per class used during training and 1000 examples per class for testing. The full non-linear model is trained with full-batch gradient descent with learning rate 0.5 and the linear model is calculated using the discretized version of Equation~\eqref{linearevol}. Test loss (accuracy) gap is calculated by averaging the loss (accuracy) over 60 runs then taking the difference between the full and the linear model. We see the predicted $1/n$ scaling with both signs. The linear model performs worse than the full model before overfitting but better after. 
(a) Training accuracy for the full model and test accuracy for both the full non-linear and linear model during training for $n=64$ and $n=1024$. We see that the performance difference between the full and linear model vanishes as $n$ gets large, and we also note that the models starts overfitting after about 2000 steps. A zoomed-in version of the test accuracy curves is shown in the inset plot.   
(b) Test accuracy gap between the full and linear model for every time step. Positive value means the full model is better than the linear. 
(c) Loss scaling with number of channels, $n$, at $t=280$ in pink and at $t=10^4$ in light blue. 
(d) Test loss gap between the full and linear model. Negative gap value means the the full model is better than the linear model. Dashed lines in pink and light blue correspond to the times the scaling is measured in subfigure (c). 
}
\label{fig:loss_panel}
\end{figure*}

We compare the training dynamics of a non-linear model and its linearized counterpart. We find that if both models are stopped at the optimal early stopping time for the non-linear model, then the non-linear model outperforms the linear model. This gap in performance fits well with the predicted $\mathcal{O}(n^{-1})$ scaling. However this does not represent a true performance gap between linear and non-linear model, but rather the fact that the linear and non-linear models achieve their maximum accuracy at different times. Indeed for late times we see the non-linear model under-performs the linear model, again with a gap scaling as $\mathcal{O}(n^{-1})$. As both linear and non-linear models over-fit, there is also a gap between the optimally stopped non-linear model and the infinite time predictions of the linear model. These results are summarized in Figure~\ref{fig:loss_panel}.

\section{Discussion}

We have presented a simple relation, Conjecture~\ref{conj:main}, for the asymptotic scaling of correlation functions with width and tested the predictions in a variety of CNN architectures.
We used these scaling relations to study the training dynamics and performance of wide convolutional networks. At infinite width, CNNs evolve as linear models, with training controlled by the constant NTK. Our conjecture gives an asymptotically tight bound on the approach to constancy.

Away from infinite width the NTK is no longer constant, but the dynamics can still be systematically approximated in a large width expansion. We use this to predict an asymptotic $\mathcal{O}(n^{-1})$ scaling for the difference between the linearized loss and the full non-linear loss \Eq{loss_diff}. In Section~\ref{sec:exp_loss} we presented evidence corroborating this prediction. 
Though the linearized and non-linear loss approach each other at infinite width, we found that their order can depend on the training time for which they are being compared. This same sensitivity is reflected in the accuracy.
In particular, depending on the stopping criteria, the full finite width model can be either better or worse then the linear model.

One motivation for extending the analysis of \cite{dyer2020asymptotics} to convolutional networks is to bridge the gap between our analytic understanding at infinite width and the empirically best performing finite width networks used in practice. Our analysis of the performance gap between linearized and non-linear models is a step in this direction. We hope the tools developed here can be extended to increasingly realistic scenarios.

\section*{Acknowledgements}
The authors wish to thank
Yasaman Bahri,
Guy Gur-Ari,
Jaehoon Lee,
Aitor Lewkowycz,
Sam Schoenholz,
and
Jascha Sohl-dickstein
for useful discussions during the completion of this work.

\bibliography{references}
\bibliographystyle{unsrt}

\appendix
\onecolumn
\section{Expansion of deep-linear network maps}
In this section, we complete the proof of Lemma~\ref{lemma:decomp}. In the body we discussed convolution layers. Here we extend the analysis to skip and global average pooling layers.
\paragraph{Skip connections}
\*
\\

\noindent Let $f(x)$ be a deep-linear network of depth $d$ with a skip connection at layer $\ell+1$. We can write
\begin{align}\label{eq:skip_decomp}
    f(x)=g(\tilde{\alpha}^{(\ell+1)}) \,=\, g(\tilde{\alpha}^{(\ell+1)}_{\textrm{straight}})+g(\alpha^{(\ell+1-k)})\,=\,\sum_{I=1}^{2}f_{I}(x)\,.
\end{align}
Here $f_{1}(x)$ and $f_{2}(x)$ are depth $d$ and $d-k$ networks with the skip connection at layer $\ell+1$ in $f(x)$ absent.
\paragraph{Global average pooling}
\*
\\

\noindent Let $f(x)$ be a deep-linear network with a global average pooling layer at depth $\ell+1$. We can write
\begin{align}\label{eq:gap_decomp}
    f(x) &= g(\tilde{\alpha}^{(\ell+1)}_{i})\,=\, \frac{1}{\mathcal{W} \mathcal{H}}\sum_{r=1}^{\mathcal{W}}\sum_{s=1}^{\mathcal{H}}g(\alpha_{r,s;i}^{(\ell)})
    \,=\,\sum_{I=1}^{W\times H} f_{I}(x)\,, \ \ \ I=\{r,s\}.
\end{align}
Here each $f_{I}(x)$ has the topology of a deep network without a the global average pooling layer.
\begin{proof}
Lemma~\ref{lemma:decomp} follows from repeated applications of \eqref{eq:conv_decomp}, \eqref{eq:skip_decomp}, and \eqref{eq:gap_decomp} to each such layer, and relabeling of indices so that $I$ runs over all terms in the resulting sums.
\end{proof}
\section{Feynman diagrams for deep-linear networks}
We are interested in computing correlation functions (Definition~\ref{def:corr}). This involves computing expectation values over the multi-variate Gaussian initial weight values. A central tool in computing these expectation values is Isserlis’ Theorem (sometimes referred to as Wick's theorem). Which establishes that higher moments under a Gaussian distribution can be computed as sums of products of second moments. For example,
\begin{align}\label{eq:isserlisexample}
    \lexpp{x} x_{1}x_{2}x_{3}x_{4}\rexp =
    \lexpp{x} x_{1}x_{2}\rexp \lexpp{x} x_{3}x_{4}\rexp
    +\lexpp{x} x_{1}x_{3}\rexp \lexpp{x} x_{2}x_{4}\rexp
    +\lexpp{x} x_{1}x_{4}\rexp \lexpp{x} x_{2}x_{3}\rexp
\end{align}

In general, keeping track of all the moments be cumbersome and Feynman diagrams are a useful book keeping tool. These diagrams were used in \cite{dyer2020asymptotics} to compute the ensemble averages over initial weights for fully-connected networks. We begin by reviewing the key definitions and results and then extend the technology as is needed to prove Theorem~\ref{thm:main}.

\subsection{Review}

Consider a deep linear network
\begin{align}
    f(x)=n^{-d/2}V^{T}W^{(d-1)}\cdots W^{(1)}Ux
\end{align}

We review how Feynman diagrams can be used to compute the asymptotic scaling of correlation functions for deep-linear networks. First we introduce some additional notation. For a factor of the network map with $\ell$ derivatives we write
\begin{align}\label{eq:derivtensor}
    T_{\mu_{1}\ldots\mu_{\ell}}(x):=\frac{\partial^{\ell}f(x)}{\partial\theta_{\mu_{1}}\cdots\partial\theta_{\mu_{\ell}}}\,.
\end{align}
We refer to $T$ as a derivative tensor. As above, if two derivative tensors in a correlation function, $C$, have $k$ paired summed indices, we say that the tensors are \emph{contracted} $k$-times in $C$.

\begin{figure*}[t!]
\centering
\captionsetup[subfigure]{oneside,margin={0.2cm,0cm}}
\subfloat[$\lexpp{\theta}f(x_{1})f(x_{2})\rexp$]{\label{fig:f2_sl}
    \includegraphics[scale=0.4, valign=c]{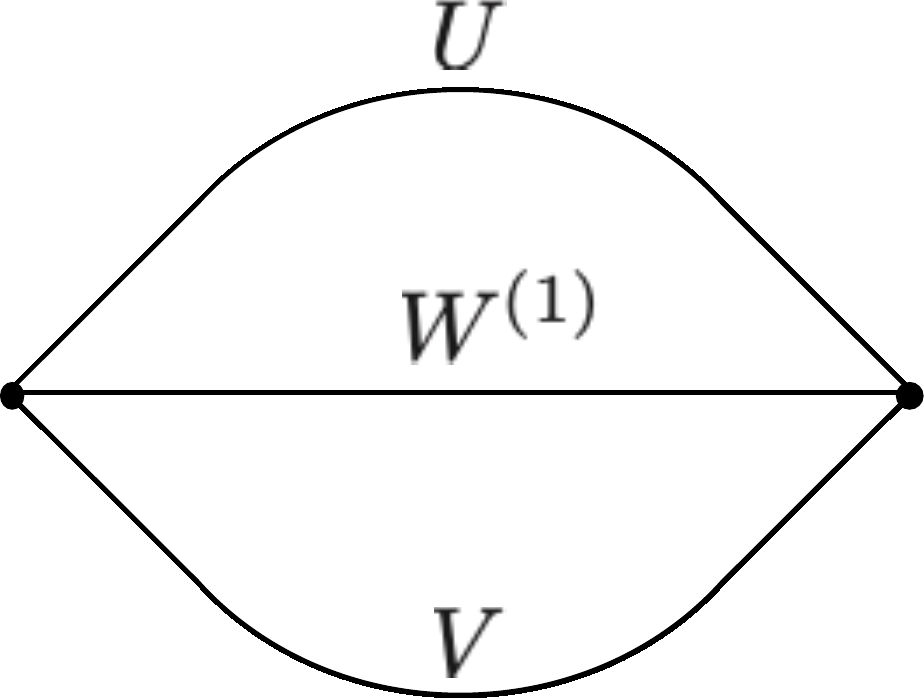}
    \vphantom{\includegraphics[scale=0.3, valign=c]{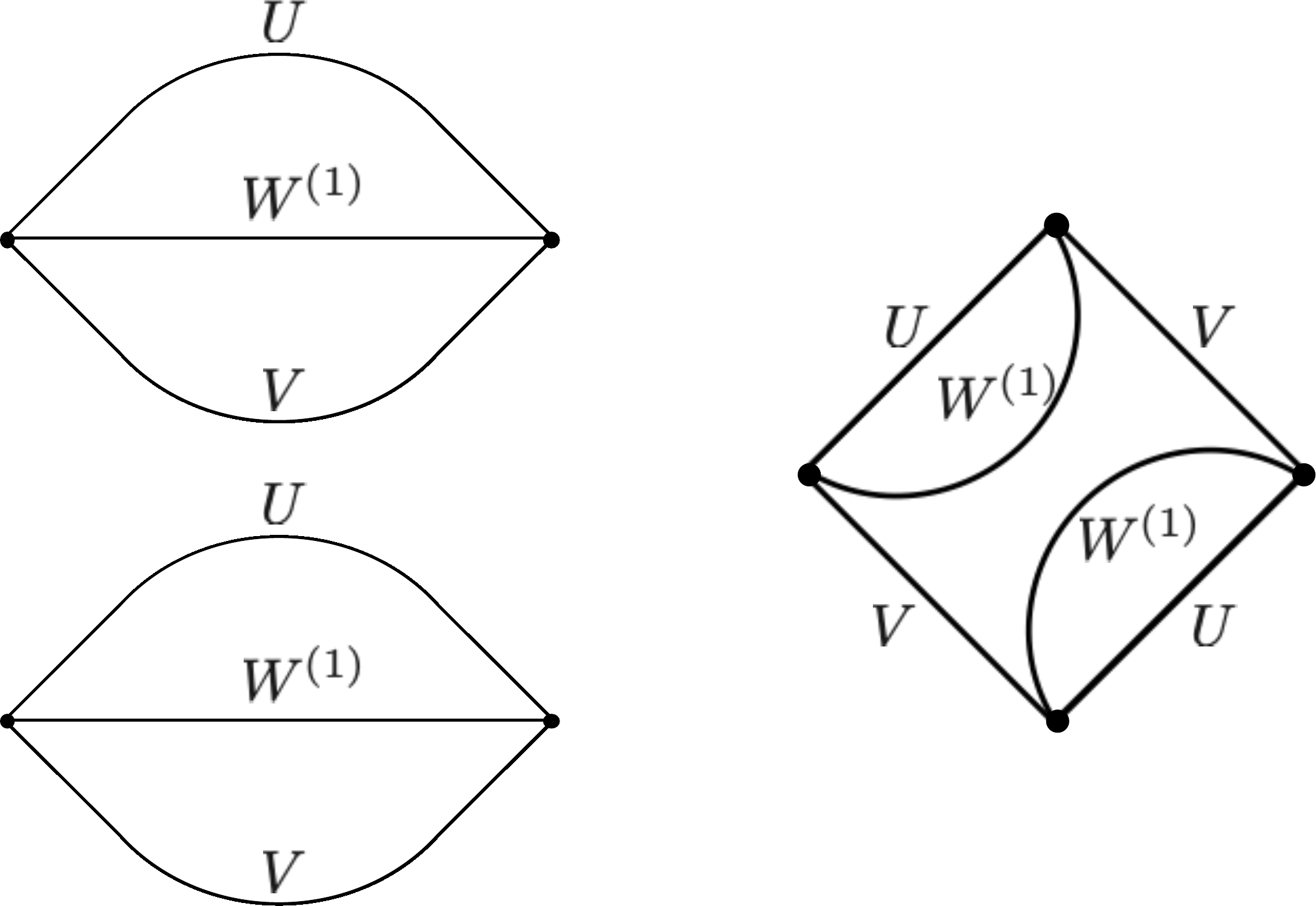}}}
\qquad\qquad
\subfloat[$\lexpp{\theta}f(x_{1})f(x_{2})f(x_{3})f(x_{4})\rexp$]{\label{fig:f4_sl}
    \includegraphics[scale=0.28, valign=c]{figs/f4_sl.pdf}}
\caption{Example Feynman diagrams for a two-hidden-layer deep-linear network. (a) Single diagram representing $\lexpp{\theta}f(x_{1})f(x_{2})\rexp$. (b) Two diagrams corresponding to $\lexpp{\theta}f(x_{1})f(x_{2})f(x_{3})f(x_{4})\rexp$.}
\label{fig:sl_diagrams}
\end{figure*}

We now describe the Feynman diagrams associated to a correlation function.
\begin{definition}\label{def:feynman}
  Let $C(x_1,\dots,x_m)$ be a correlation function for a network with $d$ hidden layers.
  The family $\Gamma(C)$ is the set of all graphs that have the following properties.
  \begin{enumerate}
  \item There are $m$ vertices $v_1,\dots,v_m$, each of degree $d+1$.
  \item Each edge has a type $t \in \{ U,W^{(1)},\dots,W^{(d-1)},V \}$. Every vertex has one edge of each type.
  \item If two derivative tensors $T_{\mu_1,\dots,\mu_\ell}(x_i),T_{\nu_1,\dots,\nu_{\ell'}}(x_j)$ are contracted $k$ times in $C$, the graph must have at least $k$ edges (of any type) connecting the vertices $v_i,v_j$.
  \end{enumerate}
  The graphs in $\Gamma(C)$ are called the \emph{Feynman diagrams} of $C$.
\end{definition}
Some example Feynman diagrams are shown in Figure~\ref{fig:sl_diagrams}.
For one hidden layer networks, the Feynman diagrams allow one to easily compute the scaling of a correlation function. Deep networks require additional technology, the double line graph.

\begin{definition} \label{def:dl}
  Let $\gamma \in \Gamma(C)$ be a Feynman diagram for a correlation function $C$ involving $k$ derivative tensors for a network of depth $d$.
  Its \emph{double-line graph}, $\DL(\gamma)$ is a graph with $kd$ vertices of degree 2, defined by the following blow-up procedure.
  \begin{itemize}
  \item Each vertex $v_i$ in $\gamma$ is mapped to $d$ vertices $v^{(1)}_i,\dots,v^{(d)}_{i}$ in $\DL(\gamma)$.
  \item Each edge $(v_{i},v_{j})$ in $\gamma$ of type $U$ is mapped to a single edge $(v^{(1)}_i,v^{(1)}_j)$.
  \item Each edge $(v_{i},v_{j})$ in $\gamma$ of type $W^{(l)}$ is mapped to two edges $(v^{(l)}_i,v^{(l)}_j)$, $(v^{(l+1)}_{i},v^{(l+1)}_{j})$.
  \item Each edge $(v_{i},v_{j})$ in $\gamma$ of type $V$ is mapped to a single edge $(v^{(d)}_i,v^{(d)}_j)$.
  \end{itemize}
  The number of \emph{faces} in $\gamma$ is given by the number of loops in the double-line graph $\DL(\gamma)$.
\end{definition}
\begin{figure*}[t!]
\centering
\captionsetup[subfigure]{oneside,margin={1.2cm,0cm}}
\subfloat[$\lexpp{\theta}f(x_{1})f(x_{2})\rexp$]{\label{fig:f2_dl}
    \includegraphics[scale=0.4, valign=c]{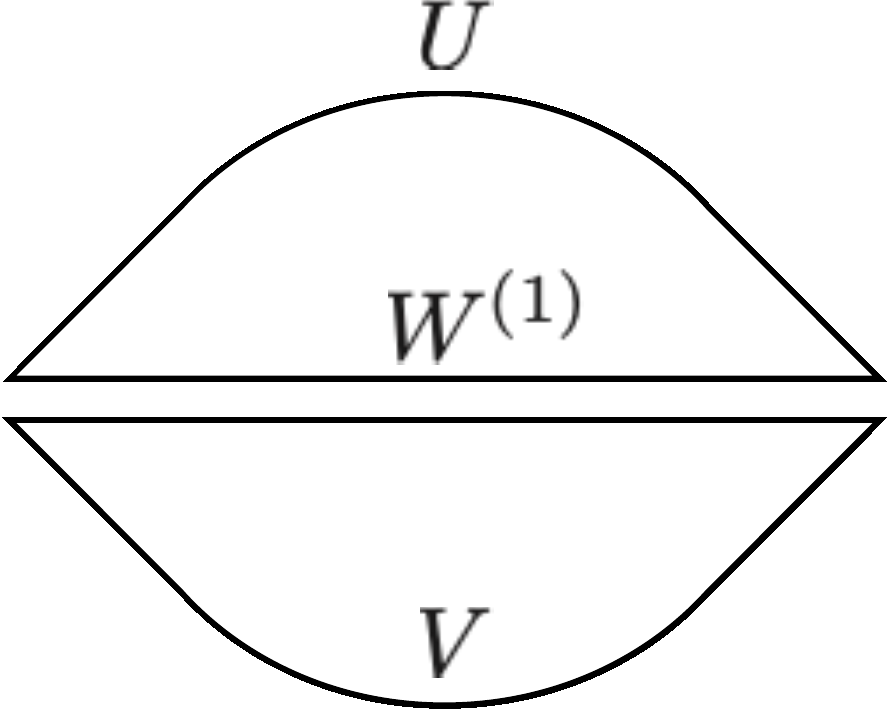}
    \vphantom{\includegraphics[scale=0.3, valign=c]{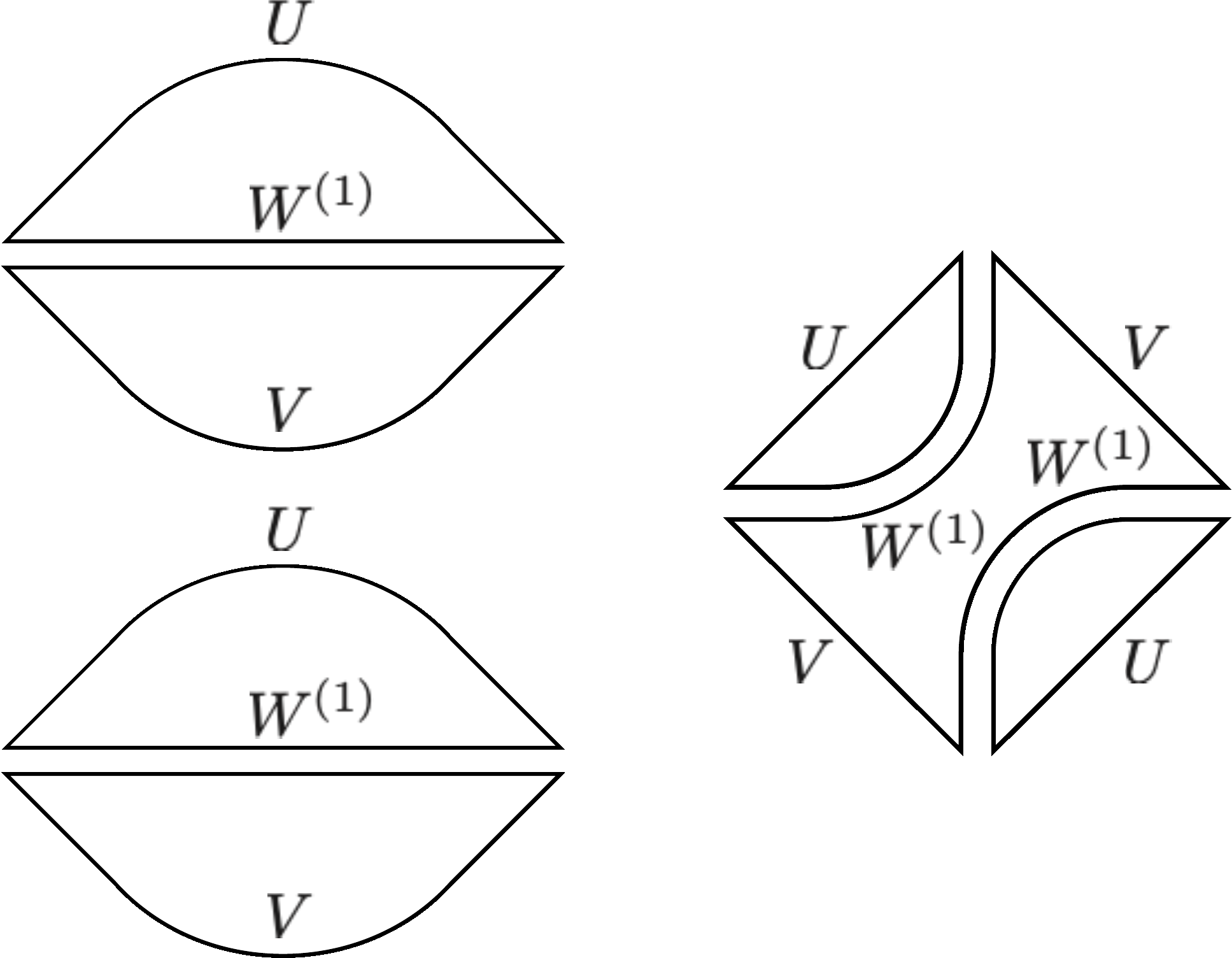}}}
\qquad\qquad\qquad\subfloat[$\lexpp{\theta}f(x_{1})f(x_{2})f(x_{3})f(x_{4})\rexp$]{\label{fig:f4_dl}
    \includegraphics[scale=0.3, valign=c]{figs/f4_dl.pdf}}
\caption{Example double line diagrams for two-hidden-layer deep-linear network. (a) Single diagram representing $\lexpp{\theta}f(x_{1})f(x_{2})\rexp$. (b) Two diagrams corresponding to $\lexpp{\theta}f(x_{1})f(x_{2})f(x_{3})f(x_{4})\rexp$.}
\label{fig:dl_diagrams}
\end{figure*}
Some example double line graphs are shown in Figure~\ref{fig:dl_diagrams}. 
With these definitions correlation functions of deep linear networks satisfy a theorem originally due to \cite{tHooft:1973alw}.
\begin{thm}\label{thm:feynDL}
  Let $C(x_1,\dots,x_m)$ be a correlation function of a deep linear network with $d$ hidden layers, and let $\gamma \in \Gamma(C)$ be a Feynman diagram.
  The diagram represents a subset of terms that contribute to $C$, and its asymptotic behavior is determined by the Feynman rules: the subset is $\mathcal{O}(n^{s_\gamma})$ where $s_\gamma = l_\gamma - \frac{dm}{2}$, and $l_\gamma$ is the number of loops in the double-line diagram $\DL(\gamma)$.
 Furthermore, the correlation function is $C = \mathcal{O}(n^s)$, where $s = \max_{\gamma\in\Gamma(C)} s_\gamma$.
\end{thm}
Below we generalize this construction to accommodate networks with convolution, skip, and GAP layers.

\subsection{Extension}
\begin{figure*}[t!]
\centering
\captionsetup[subfigure]{oneside,margin={1.2cm,0cm}}
\subfloat[$\gamma$]{\label{fig:mixed_sl}
    \includegraphics[scale=0.5]{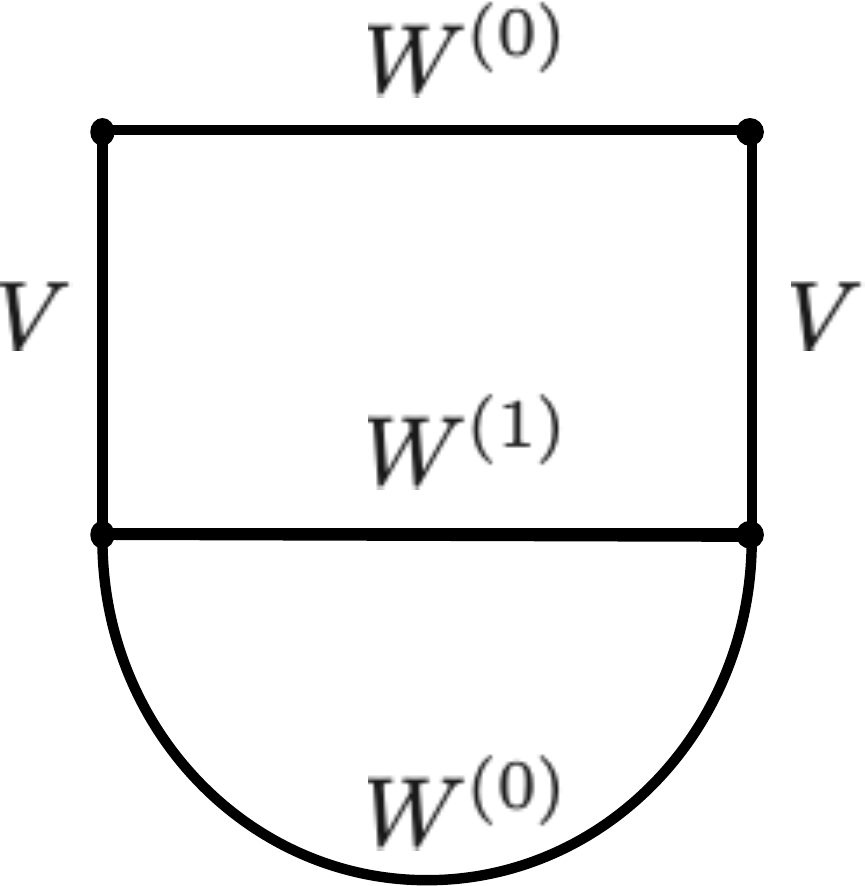}}
\qquad\qquad\qquad\subfloat[$\DL(\gamma)$]{\label{fig:mixed_dl}
    \includegraphics[scale=0.5]{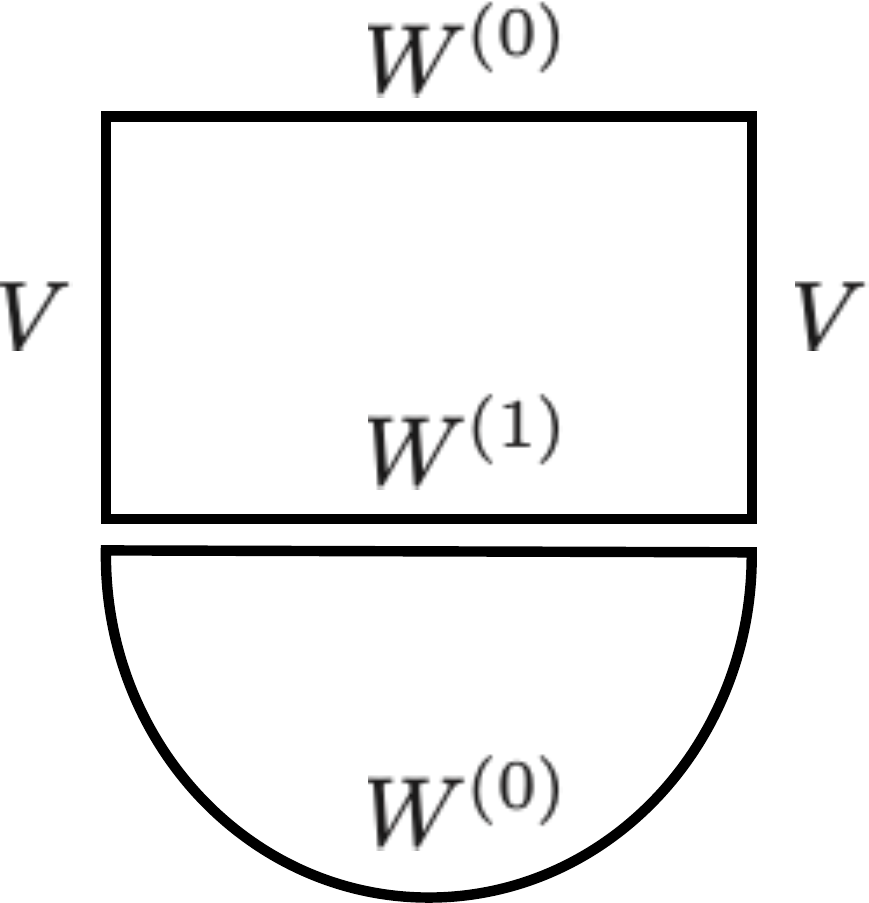}}
\caption{Example (a) single line and (b) double line diagrams for a mixed correlation function $\lexpp{\theta}f_{I_{1}}(x_2)f_{I_{2}}(x_2)f_{I_{3}}(x_3)f_{I_{4}}(x_4)\rexp$, where $f_{I_{1}}$ and $f_{I_{2}}$ have the topology of a one-hidden-layer network and $f_{I_{3}}$ and $f_{I_{4}}$ have the topology of a two-hidden-layer network. Vertices are numbered from the top left clockwise.}
\label{fig:mixed_diagrams}
\end{figure*}

We must extend the above technology to our mixed correlation functions (Definition~\ref{def:corr_mixed}).
Mixed correlation functions are expectations of maps of the form
\begin{align}
    f_{I}(x)=n^{-d_{I}/2}V^{T}_{\alpha^{I}_{d_{I}}}W_{\alpha^{I}_{d_{I}-1}}^{(\beta^{I}_{d_{i}-1})}\cdots W_{\alpha^{I}_{1}}^{(\beta^{I}_{1})}W_{\alpha^{I}_{0}}^{(0)}x\,.
\end{align}
Here the $\beta^{I}_{l}$ take values in $\{1,\ldots, d-1\}$. The $\alpha_m^{I}$ take values in the kernel indices, $\{a,b\}$. With this
$V_{\alpha}$
is a vector in $\mathbb{R}^{n}$,
$W_{\alpha}^{(\beta)} \in \mathbb{R}^{n}\times\mathbb{R}^{n}$,
and
$W_{\alpha}^{(0)} \in \mathbb{R}^{n}\times\mathbb{R}^{c_{\textrm{in}}}$,
with $c_{\textrm{in}}$ the number of input channels.
With this, we can introduce the Feynman diagrams associated to a mixed correlation function.
\begin{definition}\label{def:feynman_sl_general}
  Let $C_{I_{1},\ldots,I_{m}}(x_1,\ldots,x_m)$ be a mixed correlation function for a collection of maps, $\{f_{I_{1}},\ldots,f_{I_{m}}\}$ with depths $\{d_{1},\ldots,d_{m}\}$.
  The family $\Gamma(C_{I_{1},\ldots,I_{m}})$ is the set of all graphs that have the following properties.
  \begin{enumerate}
  \item There are $m$ vertices $v_1,\dots,v_m$, where each $v_i$ has degree $k_i=d_{i}+1$.
  \item Each edge has a type $t$. Every vertex, $v_i$ has one edge of each type $t \in \{V_{\alpha^{I_{i}}_{d_{I_{i}}}},W_{\alpha^{I}_{d_{I}-1}}^{(\beta^{I}_{d_{i}-1})},\ldots, W_{\alpha^{I}_{1}}^{(\beta^{I}_{1})},W_{\alpha^{I}_{0}}^{(0)}\}$.
  \item If two derivative tensors $T_{I_{i};\mu_1,\dots,\mu_\ell}(x_i),T_{I_{j};\nu_1,\dots,\nu_{\ell'}}(x_j)$ are contracted $s$ times in $C_{I_{1},\ldots,I_{m}}$, the graph must have at least $s$ edges (of any type) connecting the vertices $v_i,v_j$.
  \end{enumerate}
\end{definition}
 Here, we have introduced a generalized derivative tensor, $T_{I_{i};\mu_1,\dots,\mu_\ell}(x_i)=\frac{\partial^{\ell}f_{I}(x_i)}{\partial\theta_{\mu_1}\dots\partial\theta_{\mu_\ell}}$. We can again define the associated double line diagram, $\DL(\gamma)$.
\begin{definition} \label{def:dlgeneral}
  Let $\gamma \in \Gamma(C_{I_{1},\ldots,I_{m}})$ be a Feynman diagram for a mixed correlation function $C_{I_{1},\ldots,I_{m}}$ involving $m$ derivative tensors.
  Its \emph{double-line graph}, $\DL(\gamma)$ is a graph with vertices of degree 2, defined by the following blow-up procedure.
  \begin{itemize}
  \item Each vertex $v_{i}$ in $\gamma$ of degree $k_i$ is mapped to $k_i-1$ vertices $v^{(1)}_i,\dots,v^{({k_i}-1)}_{i}$ in $\DL(\gamma)$.
  \item Each edge $(v_{i},v_{j})$ in $\gamma$ of type $W_{\alpha}^{(\beta)}$ is mapped to two edges, $(v^{(e_{i})}_{i},v^{(e_{j})}_{j})$ and $(v^{(e_{i}+1)}_{i},v^{(e_{j}+1)}_{j})$.
  \item Each edge $(v_{i},v_{j})$ in $\gamma$ of type $W_{\alpha}^{(0)}$ is mapped to a single edge $(v^{(1)}_i,v^{(1)}_j)$.
  \item Each edge $(v_{i},v_{j})$ in $\gamma$ of type $V_{\alpha}$ is mapped to a single edge $(v^{(k_{i}-1)}_{i},v^{(k_{j}-1)}_{j})$.
  \end{itemize}
  Here, $e_{i}$ take values in $\{2,\ldots,k_{i}-2\}$. As in Definition~\ref{def:dl}, the number of \emph{faces} in $\gamma$ is given by the number of loops in the double-line graph $\DL(\gamma)$.
\end{definition}
Example Feynman and double-line diagrams for mixed correlation functions are shown in Figure~\ref{fig:mixed_diagrams}.

With the double line diagram defined, we can generalize Theorem \ref{thm:feynDL} for the mixed correlation functions.  
\begin{thm}\label{thm:feynDLgeneral}
  Let $C_{I_{1},\ldots,I_{m}}(x_1,\dots,x_m)$ be a mixed correlation function. Let $\gamma \in \Gamma(C_{I_{1},\ldots,I_{m}})$ be a Feynman diagram.
  The diagram represents a subset of terms that contribute to $C_{I_{1},\ldots,I_{m}}$, and its asymptotic behavior is determined by the Feynman rules: the subset is $\mathcal{O}(n^{s_\gamma})$ where $s_\gamma = l_\gamma - \sum_{i=1}^{m}\frac{k_{i}-1}{2}$, with $k_{i}$ the degree of the $i$-th vertex, and $l_\gamma$ the number of loops in the double-line diagram $\DL(\gamma)$.
 Furthermore, the mixed correlation function satisfies $C_{I_{1},\ldots,I_{m}} = \mathcal{O}(n^s)$, where $s = \max_{\gamma\in\Gamma(C_{I_{1},\ldots,I_{m}})} s_\gamma$.
\end{thm}

We now use the Feynman rules (Theorem~\ref{thm:feynDLgeneral}) to bound the scaling of a correlation function by the maximal number of connected components appearing in any single line Feynman diagram. 
\begin{lemma}\label{lemma:asympBoundgeneral}
  Let $C_{I_{1},\ldots,I_{m}}(x_1,\dots,x_m)$ be a mixed correlation function.
  Let $c_\gamma$ be the number of connected components of a graph $\gamma \in \Gamma(C_{I_{1},\ldots,I_{m}})$.
  Then $C_{I_{1},\ldots,I_{m}} = \mathcal{O}(n^s)$, where
  \begin{align}
    s = \max_{\gamma \in \Gamma(C_{I_{1},\ldots,I_{m}})} c_\gamma - \frac{m}{2} \,. \label{eq:dl_bound_totalwbias}
  \end{align}
\end{lemma}
\begin{proof}
It is enough to show that each connected component $\gamma'$ in $\gamma$ is bounded as $\mathcal{O}(n^{s_{\gamma'}})$ where $s_{\gamma'}\leq 1-\frac{v_{\gamma'}}{2}$. The graph $\DL(\gamma')$ is a triangulation of a Riemann surface with $f$ faces $e$ edges, and $v_{\gamma'}$ vertices of degrees $k_{1},\ldots,k_{v_{\gamma'}}$. The Feynman rules give
\begin{align}
 s_{\gamma'}=f-\sum_{i=1}^{v_{\gamma'}}\frac{k_{i}-1}{2}\,. \label{eq:feyn_dscaling_wbias}
\end{align}
Using the relation $e=\sum_{i=1}^{v_{\gamma'}}\frac{k_{i}}{2}$ and the definition of the Euler character, $\chi=v_{\gamma'}-e+f$ we have
\begin{align}
 s_{\gamma'}=\chi-\frac{v_{\gamma'}}{2}\,. \label{eq:feyn_dscaling_euler_wbias}
\end{align}
The diagram $\DL(\gamma')$ is a triangulation of a Riemann surface with at least one boundary, thus $\chi\leq 1$ and $s_{\gamma'}\leq 1-\frac{v_{\gamma'}}{2}$. 
\end{proof}
We are now ready to prove Theorem~\ref{thm:main} for deep-linear networks. 
\begin{proof}
Let $C(x_{1},\ldots,x_{m})$ be a correlation function for a deep-linear network built out of dense, convolution, skip, and GAP layers. By Lemma~\ref{lemma:decomp} we can write
\begin{align}
    C(x_{1},\ldots,x_{m})=\sum_{I_{1},\ldots,I_{m}}C_{I_{1},\ldots,I_{m}}(x_{1},\ldots,x_{m})\,.
\end{align}
Thus, by Lemma~\ref{lemma:asympBoundgeneral} $C(x_{1},\ldots,x_{m})=\mathcal{O}(n^{s})$ where
\begin{align}
    s=\max_{\gamma\in\Gamma(C_{I_{1},\ldots,I_{m}})} c_{\gamma}-\frac{m}{2}
\end{align}
We now show that $c_{\gamma}\leq n_e+\frac{n_o}{2} \ \forall \, \gamma$. Firstly note that the cluster graph $G_{C}$ (Definition~\ref{def:clust}) is a sub-graph of all $\gamma\in\Gamma(C_{I_{1},\ldots,I_{m}})$, thus $c_{\gamma}\leq n_e+n_o$ as each cluster in $G_{C}$ can form at most one connected component in $\gamma$. Furthermore, note that each connected component in $\gamma$ contains an even number of vertices, thus even clusters in $G_{C}$ can form there own connected components in $\gamma$, but odd clusters must be paired in the connected components of $\gamma$. Thus, $c_{\gamma}\leq n_e+\frac{n_o}{2}$.
\end{proof}

\section{One-hidden-layer non-linear Networks}

In this section we prove Theorem~\ref{thm:main} for the case of networks with a single hidden layer. In this case, there are no skip connections, so we consider networks with a single convolutional layer terminated by either a gap or flatten layer.
\begin{align}
    f_{\textrm{Flatten}}(x)&=\frac{1}{\sqrt{\mathcal{W} \mathcal{H} n}}\sum_{r=1}^{\mathcal{W}}\sum_{s=1}^{\mathcal{H}}\sum_{i=1}^{n}V_{r,s;i}\alpha^{(1)}_{r,s;i}\,, \\ 
f_{\textrm{GAP}}(x)\,&=\,\frac{1}{\mathcal{W} \mathcal{H}\sqrt{n}}\sum_{r=1}^{\mathcal{W}}\sum_{s=1}^{\mathcal{H}}\sum_{i=1}^{n}V_{i}\alpha^{(1)}_{r,s;i}\,,\nonumber\\
\alpha^{(1)}_{r,s;i}&=\sigma\left(\frac{1}{\sqrt{k_{w}k_{h}c_{\textrm{in}}}}\sum_{a=1}^{k_{w}}\sum_{b=1}^{k_{h}}\sum_{j=1}^{c_{\textrm{in}}}W^{(0)}_{a,b;ij}x_{r+a,s+b;j}\right)\,.
\end{align}

It is convenient to adopt a notation that highlights the scaling with respect to the network width (number of convolutional channels). To this end, we write the network function as,
\begin{align}\label{eq:simp_ohl}
    f(x)&=\frac{1}{\sqrt{n}}\sum_{i=1}^{n}\mathbf{V}_{i}\sigma(\mathbf{X}\mathbf{U}_{i})\,.
\end{align}
To write $f$ in this form, we have juggled the indexing over input pixels, channels, and kernels as follows.
\begin{itemize}
    \item $\mathbf{X}$ is a  $\mathcal{W}\mathcal{H} \times k_{w} k_{h} c_{\textrm{in}}$ matrix. $\mathbf{X}_{\{r,s\},\{a,b,j\}}:=x_{r+a,s+b;j}$.
    \item For each $i$, $\mathbf{U}_{i}$ is a $k_{w} k_{h} c_{\textrm{in}}$ vector. $\left(\mathbf{U}_{i}\right)_{\{a,b,j\}}:=W^{(0)}_{a,b;ij}$.
    \item For each $i$, $\mathbf{V}_{i}$ is a $\mathcal{W}\mathcal{H}$ vector. For flatten models, $\left(\mathbf{V}_{i}\right)_{\{r,s\}}=V_{r,s;i}$, while for GAP models $\left(\mathbf{V}_{i}\right)_{\{r,s\}}=V_{i}\delta_{rs}$.
\end{itemize}
Here $\delta_{rs}$ is the Kronecker delta and $c_{\textrm{in}}$ is the number of input channels. We have also dropped the normalization over input channel number, kernel size, and image size as they do not effect the scaling with respect to width. 

If we adopt the notation $R=\{r,s\}$, $A=\{a,b,j\}$, and let $\mathbf{U}_{A,i}$ and $\mathbf{V}_{i,R}$ be Gaussian distributed with unit variance. 
\begin{align}\label{eq:gaussian}
    \lexpp{\theta}\mathbf{U}_{A,i}\mathbf{U}_{B,j}\rexp\,=\,\delta_{AB}\delta_{ij}\,, \ \ \ \lexpp{\theta}\mathbf{V}_{i,R}\mathbf{V}_{j,S}\rexp\,=\,\delta_{ij}\delta_{RS}\,.
\end{align}

With all of this notation out of the way, we can prove Theorem~\ref{thm:main}. An outline of the argument is the following. A general correlation function can be written as a sum of terms,
\begin{align}
    &C(x_{1},\ldots,x_{m})=n^{-m/2}\sum_{\alpha=1}^{K}\sum_{i_{1},\ldots,i_{r_{\alpha}}=1}^{n}\mathcal{S}^{(\alpha)}_{i_{1}\ldots i_{r_{\alpha}}}\label{eq:alpha_sum}\\
    &\mathcal{S}^{(\alpha)}_{i_{1}\ldots i_{r_{\alpha}}}=
    \sum_{A_{1},\ldots ,A_{r_{\alpha}}=1}^{k_w k_h c_{\textrm{in}}}
    \sum_{R_{1},\ldots ,R_{r_{\alpha}}=1}^{\mathcal{W}\mathcal{H}}
    \mathcal{M}_{A_{\kappa^{\alpha}_{1}}\ldots A_{\kappa^{\alpha}_{m}},R_{\kappa^{\alpha}_{1}}\ldots R_{\kappa^{\alpha}_{m}}}(x_1, \ldots, x_m) \label{eq:S_alpha} \\ &\qquad\qquad\qquad \times \lexpp{
    \mathbf{U}}
    \sigma^{(\ell_{1})}(\mathbf{X}_{1;A_{\kappa^{\alpha}_{1}}R_{\kappa^{\alpha}_{1}}}\mathbf{U}_{A_{\kappa^{\alpha}_{1}}i_{\kappa^{\alpha}_{1}}})\cdots
    \sigma^{(\ell_{m})}(\mathbf{X}_{m;A_{\kappa^{\alpha}_{m}}R_{\kappa^{\alpha}_{m}}}\mathbf{U}_{A_{\kappa^{\alpha}_{m}}i_{\kappa^{\alpha}_{m}}})\rexp\,.\nonumber
\end{align}
Here, the subscripts $\{\kappa_{1}^{\alpha},\ldots\kappa^{\alpha}_{m}\}$ take values in $\{1,\ldots,r_{\alpha}\}$; the superscripts $\sigma^{(\ell)}$ indicates the $\ell$-th derivative of $\sigma$; and $\mathcal{M}$ is an $n$-independent numerical factor. Here each term in the sum over $\alpha$ has $r_{\alpha}$ \emph{index} sums, $\{i_{1},\ldots,i_{r_{\alpha}}\}$, which run over $n$ possible values.

This form of the correlation function follows from computing the expectation over the weights $\mathbf{V}$ and evaluating all derivatives in the correlation function. These operations generate a sum of $K$ expectation values over $\mathbf{U}$ and $\alpha$ indexes the terms in this sum. 
Before completing the proof, as an explicit example, let us evaluate \Eq{alpha_sum} for the NTK,

\begin{align}
    C(x_1, x_2) = \lexpp{\theta}\Theta(x_1, x_2) \rexp 
    = \lexpp{\theta} \sum_{i, R} \frac{\partial f(x_1)}{\partial \mathbf{V}_{i,R}} \frac{\partial f(x_2)}{\partial \mathbf{V}_{i,R}}\rexp + 
    \lexpp{\theta} \sum_{A, i}\frac{\partial f(x_1)}{\partial \mathbf{U}_{A, i}} \frac{\partial f(x_2)}{\partial \mathbf{U}_{A, i}}  \rexp
\end{align}
Expanding the first term,
\begin{align}
&\sum_{i, R} \lexpp{\theta} \frac{\partial f(x_1)}{\partial \mathbf{V}_{i,R}} \frac{\partial f(x_2)}{\partial \mathbf{V}_{i,R}}  \rexp \nonumber
\\&=\frac{1}{n}  \sum_{i_1, i_2 = 1}^n \sum_{A_{1} ,A_{2}=1}^{k_w k_h c_{\textrm{in}}} \sum_{R_{1} ,R_{2}=1}^{\mathcal{W}\mathcal{H}} \sum_{i, R} 
\lexpp{\mathbf{V}}\frac{\partial \mathbf{V}_{i_1, R_1}}{\partial \mathbf{V}_{i, R}}\frac{\partial \mathbf{V}_{i_2, R_2}}{\partial \mathbf{V}_{i, R}}\rexp \\ 
&\qquad\qquad\qquad\qquad\qquad\qquad\qquad\times\lexpp{\mathbf{U}}\sigma(\mathbf{X}_{1;A_1 R_{1}}\mathbf{U}_{A_1 i_{1}})\sigma(\mathbf{X}_{2;A_2 R_{2}}\mathbf{U}_{A_2 i_{2}})\rexp \nonumber\\ 
&=\frac{1}{n}  \sum_{i_1, i_2 = 1}^n \sum_{A_{1} ,A_{2}=1}^{k_w k_h c_{\textrm{in}}}\sum_{R_{1} ,R_{2}=1}^{\mathcal{W}\mathcal{H}} \sum_{i, R} \delta_{i_1 i}\delta_{i_2 i}\delta_{R_1 R}\delta_{R_2 R}\\ &\qquad\qquad\qquad\qquad\qquad\qquad\qquad\times\lexpp{\mathbf{U}}\sigma(\mathbf{X}_{1;A_1 R_{1}}\mathbf{U}_{A_1 i_{1}})\sigma(\mathbf{X}_{2;A_2 R_{2}}\mathbf{U}_{A_2 i_{2}})\rexp \nonumber\\
&=\frac{1}{n} \sum_{i}^n\sum_{A_{1} ,A_{2}=1}^{k_w k_h c_{\textrm{in}}}\sum_{R=1}^{\mathcal{W}\mathcal{H}}  \lexpp{\mathbf{U}}\sigma(\mathbf{X}_{1;A_1 R}\mathbf{U}_{A_1 i})\sigma(\mathbf{X}_{2;A_2 R}\mathbf{U}_{A_2 i})\rexp
\end{align}

Similarly for the second term,
\begin{align}
&\sum_{A, i} \lexpp{\theta} \frac{\partial f(x_1)}{\partial \mathbf{U}_{A ,i}} \frac{\partial f(x_2)}{\partial \mathbf{U}_{A,i}}  \rexp =  \\
&=\frac{1}{n}  \sum_{i_1, i_2 = 1}^n \sum_{A_{1} ,A_{2}=1}^{k_w k_h c_{\textrm{in}}}\sum_{R_{1} ,R_{2}=1}^{\mathcal{W}\mathcal{H}} \sum_{A, i} 
\delta_{i_1 i_2}\delta_{R_1 R_2}\\  
&\quad\times\lexpp{\mathbf{U}}
\sigma'(\mathbf{X}_{1; A_1 R_1}\mathbf{U}_{A_1 i_{1}})\mathbf{X}_{1;A_1 R_1}\delta_{A_1 A}\delta_{i_1 i}
\sigma'(\mathbf{X}_{2; A_2 R_2}\mathbf{U}_{A_2 i_{2}})\mathbf{X}_{2;A_2 R_2}\delta_{A_2 A}\delta_{i_2 i} \nonumber
\rexp\\
&=\frac{1}{n}  \sum_{i = 1}^n \sum_{A=1}^{k_w k_h c_{\textrm{in}}}\sum_{R=1}^{\mathcal{W}\mathcal{H}}
\mathbf{X}_{1;A R}  \mathbf{X}_{2;A R}  
\lexpp{\mathbf{U}}
\sigma'(\mathbf{X}_{1; A R}\mathbf{U}_{A i})
\sigma'(\mathbf{X}_{2; A R}\mathbf{U}_{A i})
\rexp
\end{align}

Combined, we find 
\begin{align}
    C(x_1, x_2) &= \lexpp{\theta}\Theta(x_1, x_2) \rexp \\
    &=\frac{1}{n}  \sum_{i = 1}^n \sum_{R=1}^{\mathcal{W}\mathcal{H}}
\left[ \sum_{A_{1} ,A_{2}=1}^{k_w k_h c_{\textrm{in}}} \lexpp{\mathbf{U}}\sigma(\mathbf{X}_{1;A_1 R}\mathbf{U}_{A_1 i})\sigma(\mathbf{X}_{2;A_2 R}\mathbf{U}_{A_2 i}) \rexp\right. \\
&\qquad\qquad\quad \ + \left.
\sum_{A=1}^{k_w k_h c_{\textrm{in}}}
\mathbf{X}_{1;A R}  \mathbf{X}_{2;A R}  
\lexpp{\mathbf{U}}
\sigma'(\mathbf{X}_{1; A R}\mathbf{U}_{A i})
\sigma'(\mathbf{X}_{2; A R}\mathbf{U}_{A i})
\rexp
\right]\nonumber
\end{align}

In the general case, we argue that the maximum number of sums, $r_{\textrm{max}}=\max_{\alpha}r_{\alpha}$ is bounded as $r_{\textrm{max}}\leq n_{e}+\frac{n_{o}}{2}$, where $n_{e}$ and $n_{o}$ are the number of even and odd clusters in the cluster graph $G_{C}$. We further argue that the $\mathcal{S}^{(\alpha)}_{i_{1}\ldots i_{r_{\alpha}}}$ are bounded by an $n$-independent constant. These two statements establish Theorem~\ref{thm:main}.

To prove these results we again take a graphical approach. We introduce a new class of graphs to keep track of the index sums in $C$.
\begin{definition}\label{def:ind}
Let $C(x_1,\dots,x_m)$ be a correlation function for a one-hidden-layer non-linear network.
  The family $\Gamma'(C)$ is the set of all graphs that have the following properties.
  \begin{enumerate}
  \item There are $m$ vertices $v_1,\dots,v_m$, each of degree at least one.
  \item Each edge has a type $t \in \{ U,V \}$. Every vertex has one edge of type $V$.
  \item There is an edge $(v_{i}, v_{j})$ for $f(x_{i})$, $f(x_{j})$ contracted in $C$.
  \end{enumerate}
\end{definition}
Each graph $\gamma_{\alpha}\in\Gamma'(C)$ corresponds to a term in the $\alpha$ sum \eqref{eq:alpha_sum}. The number of index sums, $r_{\alpha}$ is the number of connected components in $\gamma_{\alpha}$. To see this, note that we have one index sum for each factor of the network map $f(x)$ (Equation~\eqref{eq:simp_ohl}) in the correlation function $C$. A contracted derivative between pairs of network maps in $C$, results in a delta function eliminating one index sum. Similarly, the $V$ edges in $\gamma_{\alpha}$ correspond to using Isserlis’ theorem to evaluate the Gaussian expectation over the $\mathbf{V}_{i}$ in $C$. A single covariance as in \eqref{eq:gaussian} also eliminates an index sum due to the Kronecker delta factor. The result is that each connected component corresponds to a single sum.

We now argue for the bound on the maximal number of connected components.
\begin{lemma}
  Let $C(x_1,\dots,x_m)$ be a correlation function for a one-hidden-layer non-linear network, let $\gamma_{\alpha}\in\Gamma'(C)$, and let $c_{\alpha}$ denote the number of connected components in $\gamma_{\alpha}$. Further, let $G_{C}$ be the cluster graph of $C$ and $n_{e}$ ($n_{o}$) denote the number of even (odd) clusters. Then $c_{\alpha}$ satisfies.
  \begin{align}
      c_{\alpha}\leq n_{e}+\frac{n_{o}}{2}\,.
  \end{align}
\end{lemma}

\begin{proof}
First note that $G_{C}$ is a sub-graph of $\gamma_{\alpha}$. As each vertex in $\gamma_{\alpha}$ must have one $V$ edge, connected components in $\gamma_{\alpha}$ have an even number of vertices, thus the even clusters in $G_{C}$ can form their own connected components in $\gamma_{\alpha}$ while an odd clusters in $G_{C}$ must pair with at least one other odd cluster to form a connected component in $\gamma_{\alpha}$. 
\end{proof}

An immediate consequence of this lemma is the bound $r_{\textrm{max}}\leq n_{e}+\frac{n_{o}}{2}$. We now prove Theorem~\ref{thm:main} for one-hidden-layer non-linear networks.
\begin{proof}
We can bound the correlation function $C(x_{1},\ldots,x_{m})$ as,
\begin{align}
    C(x_{1},\ldots,x_{m})&\leq n^{-\frac{m}{2}}\sum_{\alpha=1}^{K}\sum_{i_{1},\ldots,i_{r_{\alpha}}=1}^{n}|\mathcal{S}^{(\alpha)}_{i_{1}\ldots i_{r_{\alpha}}}|\nonumber\\
    &\leq K n^{r_{\textrm{max}}-\frac{m}{2}} s_{\textrm{max}}\,\leq\, K n^{n_{e}+\frac{n_{o}}{2}-\frac{m}{2}} s_{\textrm{max}}
\end{align}
Here we have introduced $s_{\textrm{max}}$ as a bound on the expectation values, $s_{\textrm{max}}=\max_{\alpha,i_{1},\ldots,i_{r_{\alpha}}}|\mathcal{S}^{(\alpha)}_{i_{1}\ldots i_{r_{\alpha}}}|$. The expectations $S^{(\alpha)}_{i_{1}\ldots i_{r_{\alpha}}}$ can only take $\mathcal{O}(1)$ different values, as the $\mathbf{U}_{i}$ are i.i.d. $s_{\textrm{max}}$ is the maximum over these $\mathcal{O}(1)$ options. Thus $C=\mathcal{O}(n^{n_{e}+\frac{n_{o}}{2}-\frac{m}{2}})$.
\end{proof}

\section{Late Time Behavior of GAP Networks}
In this section we will briefly discuss some non-trivial features we see during training of convolutional neural networks with global average pooling. When training these models until convergence, we see a discontinuity in the NTK and the network function at late times that has to be avoided to see the theoretically predicted scaling behavior with width. 

\begin{figure}[t!]
\centering
\captionsetup[subfigure]{oneside}
\subfloat[Evolution of NTK, and training loss and accuracy for width 128]{\label{fig:late_time_behavior_a}
    \includegraphics[scale=0.38]{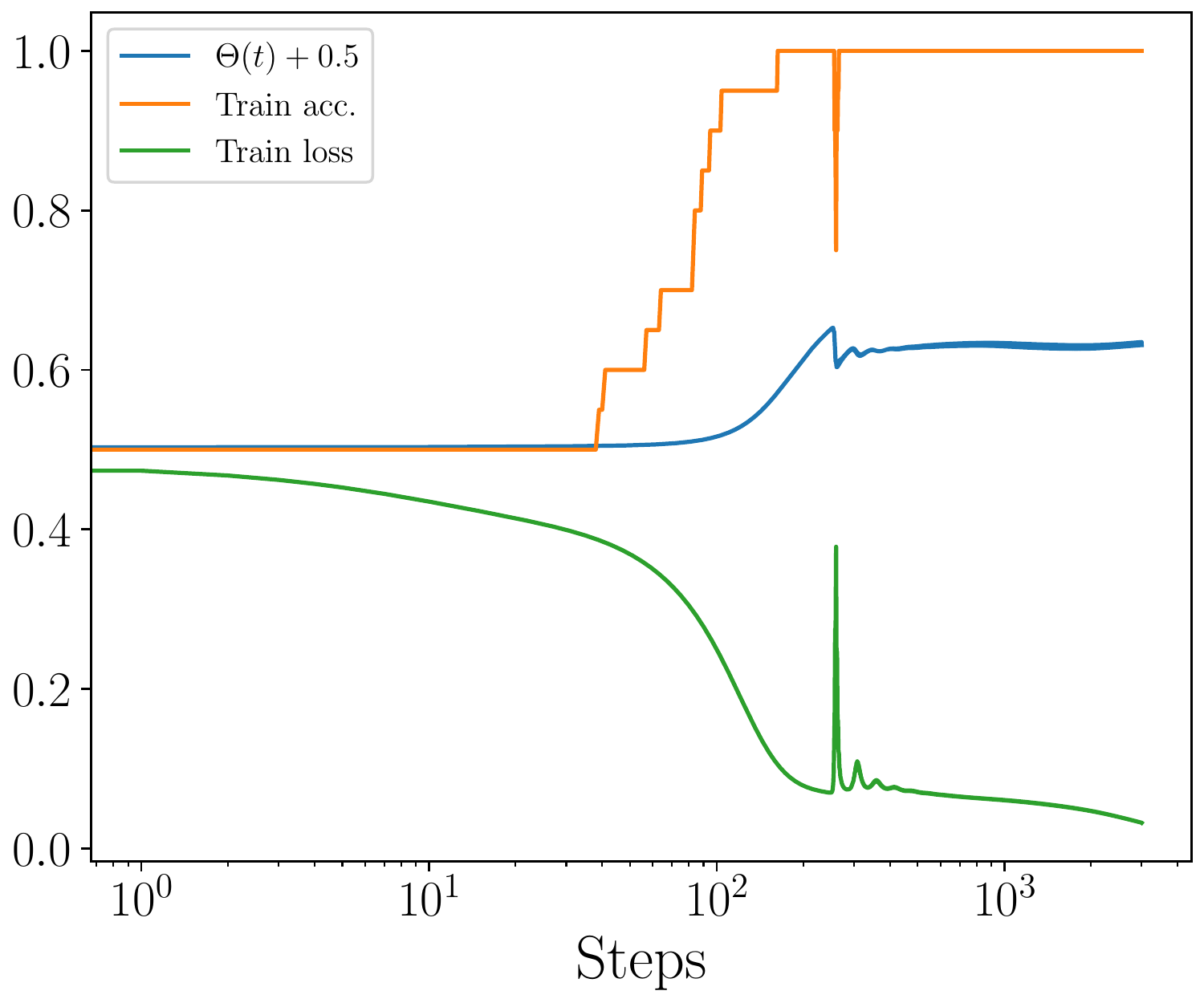}}
\subfloat[NTK evolution for different widths]{\label{fig:late_time_behavior_b}
    \includegraphics[scale=0.38]{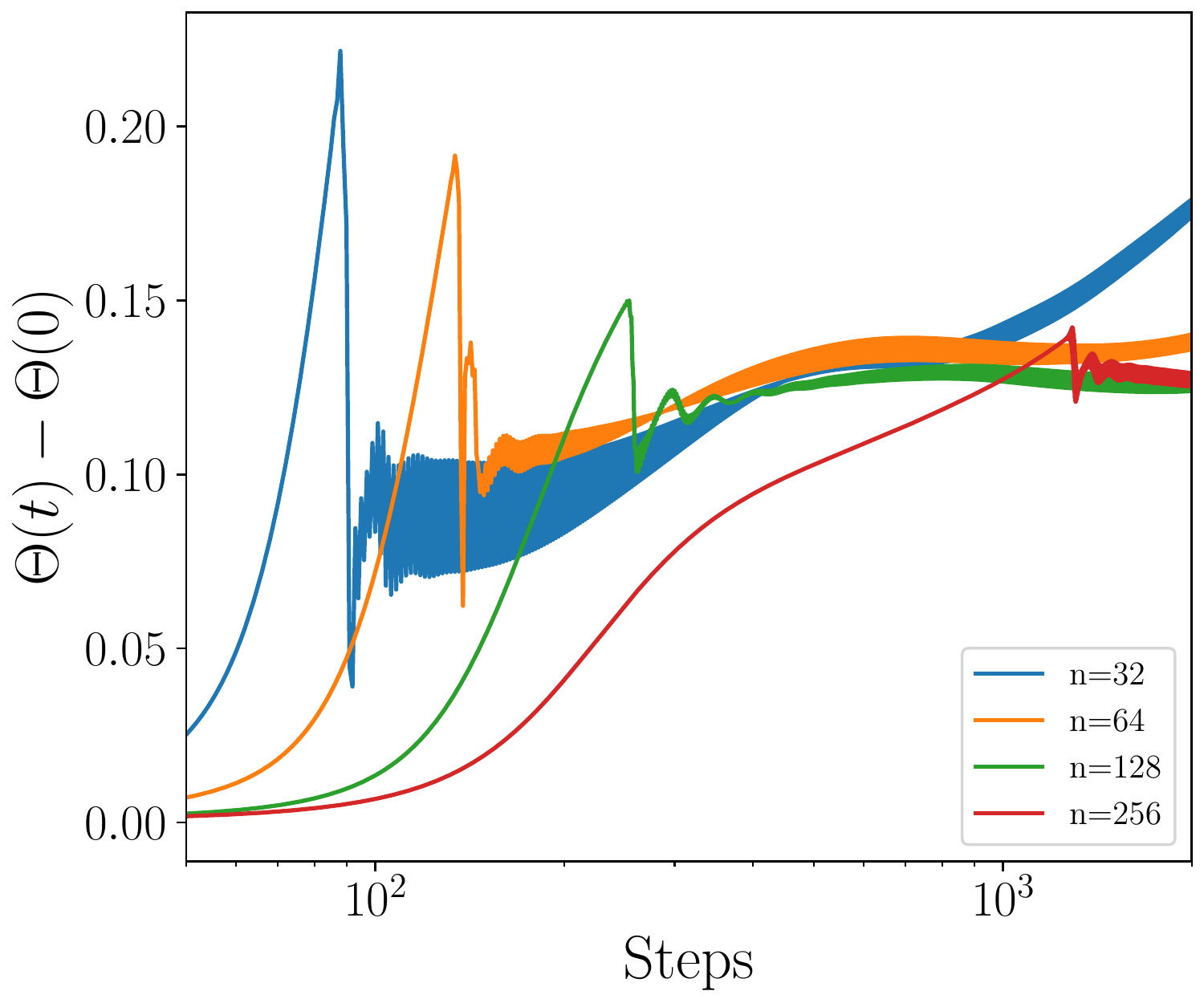}}
\caption{Evolution of a three-hidden-layer CNN width tanh activation function and global average pooling on 2-class MNIST with 10 examples per class and trained with gradient descent and learning rate 1. Late in training we see a high frequency oscillation in the NTK and large spikes in the loss and accuracy.}
\label{fig:late_time_behavior}
\end{figure}

One such example is shown in \Fig{late_time_behavior_a}, where the evolution of the NTK and the training loss and accuracy exhibits a discontinuity after about 250 steps. The evolution of the NTK for different widths is shown in  \Fig{late_time_behavior_b}, which shows the discontinuity and a high frequency oscillation that is width dependent. We also note that the effect gets smaller with larger width. The number of steps until this feature appears does depend on the hyperparameters in a systematic way, so they can be chosen such that this behavior can be avoided. It occurs earlier for deeper networks, and later for wider networks; later for smaller learning rates; and increasing the precision from float-32 to float-64 slightly delays it.

\section{Variance of NTK}
Here we establish the claim that the variance of the NTK is $\mathcal{O}(n^{-1})$ for deep-linear networks with convolutional, skip, GAP layers as well as for one-hidden-layer non-linear CNNs.

\paragraph{Deep-linear convolutional networks} In this case we can argue using Lemma~\ref{lemma:decomp} and the Feynman diagrams. 
The variance of the kernel takes the form 
\begin{align}
    \textrm{Var}_{\theta}\left[\Theta(x,x')\right]&= \sum_{I=1}^{\mathcal{N}_{f}}\textrm{Var}_{\theta}\left[\Theta_{I}(x,x')\right] \,=\, \lexpp{\theta}\Theta_{I}(x,x')^{2}\rexp-\lexpp{\theta}\Theta_{I}(x,x')\rexp^{2}
\end{align}
Here we have introduced the notation $\Theta_{I}(x,x')=\sum_{\mu}\frac{\partial f_{I}(x)}{\partial\theta_{\mu}}\frac{\partial f_{I}(x')}{\partial\theta_{\mu}}$.
This form for the variance follows from using Lemma~\ref{lemma:decomp} to write $f=\sum_{I=1}^{\mathcal{N}_{f}}f_{I}$ and noting that $\lexpp{\theta}f_{I}(x)f_{J}(x')\rexp=0$ if $I\neq J$. This last statement is a result of the fact that for $I\neq J$, the maps $f_{I}$ and $f_{J}$ necessarily contain different weights. As the weights are initialized i.i.d. with zero mean the expectation of a product of two different maps vanishes. 

We are now in a position to use the Feynman diagrams to compute the scaling of $\textrm{Var}_{\theta}\left[\Theta_{I}(x,x')\right]$. If we denote by $C_{1;I}$ the expectation of $\Theta_{I}$ and $C_{2;I}$ the expectation of the square.
\begin{align}\label{eq:varcordef}
    C_{1;I}(x,x')&=\lexpp{\theta}\Theta_{I}(x,x')\rexp\\
    C_{2;I}(x,x')&=\lexpp{\theta}\Theta_{I}(x,x')^{2}\rexp\,,
\end{align}
$C_{1}$ can be computed from the double line diagrams associated to each graph in $\Gamma(C_{1})$ and $C_{2}$ can be computed from the graphs in $\Gamma(C_{2})$. For the variance, we actually need to compute $C_{1}^{2}$. The value of $C_{1}$ is given by summing the result of applying the Feynman rules to each diagram in $\Gamma(C_{1})$. Thus $C_{1}^{2}$ is given by applying the Feynman rules to every graph in $\Gamma(C_{1})\times\Gamma(C_{1})$.

From the rules for drawing Feynman diagrams, Definition~\ref{def:feynman_sl_general}, every diagram $\gamma\in\Gamma(C_{1})\times\Gamma(C_{1})$ is also a valid diagram in $\Gamma(C_{2})$. The scaling of the variance can thus be read off by considering the double-line diagrams $\DL(\gamma)$ for $\gamma\in \Gamma(C_{2}) \setminus \Gamma(C_{1})\times\Gamma(C_{1})$. These diagrams have four vertices, but a single connected component, thus the Feynman rules (Theorem~\ref{thm:feynDLgeneral}) give $\textrm{Var}_{\theta}[(x,x')]=\mathcal{O}(n^{-1})$

\paragraph{One-hidden-layer non-linear convolutional networks} Here, we proceed by brute force calculation. Adopting the notation of Equation~\eqref{eq:simp_ohl}, the NTK can be written as
\begin{align}\label{eq:ohl_ntk}
    \Theta(x,x')=\frac{1}{n}\sum_{i=1}^{n}&\Bigg{(}\sigma\left(\mathbf{X}\mathbf{U}_{i}\right)^{T}\sigma\left(\mathbf{X}'\mathbf{U}_{i}\right)  \\
    &+\sum_{R,S,A}\mathbf{V}_{iR}\mathbf{V}_{iS}\sigma'\left(\mathbf{X}_{RA}\mathbf{U}_{Ai}\right)\mathbf{X}_{RA}
    \sigma'\left(\mathbf{X}'_{SA}\mathbf{U}_{Ai}\right)\mathbf{X'}_{SA}\Bigg{)}\,.\nonumber
\end{align}
To simplify notation, we write the above as $\Theta(x,x')=\frac{1}{n}\sum_{i=1}^{n}\phi[\mathbf{U}_{i},\mathbf{V}_{i}]$. The variance of the NTK can be written as
\begin{align}
    \textrm{Var}_{\theta}\left[\Theta(x,x')\right]&=\lexpp{\theta}\Theta(x,x')^{2}\rexp-\lexpp{\theta}\Theta(x,x')\rexp^{2}\,\\
    &=\,\frac{1}{n^{2}}\sum_{i=1}^{n}\left(\lexpp{\theta}\phi[\mathbf{U}_{i},\mathbf{V}_{i}]^{2}\rexp-\lexpp{\theta}\phi[\mathbf{U}_{i},\mathbf{V}_{i}]\rexp^{2}\right)\\
    &=\frac{1}{n}\textrm{Var}_{\theta}\left[\phi[\mathbf{U},\mathbf{V}]\right]=\mathcal{O}(n^{-1})\,.
\end{align}

In arriving at the last line we have used the fact that $\mathbf{U}_{i},\mathbf{V}_{i}$ are i.i.d. and that $\textrm{Var}_{\theta}\left[\phi[\mathbf{U},\mathbf{V}]\right]=\mathcal{O}(1)$.
\end{document}